\documentclass[runningheads]{llncs}

\usepackage{eccv}

\usepackage{eccvabbrv}

\usepackage{graphicx}
\usepackage{booktabs}
\usepackage {pifont} % \ding {xx}
\usepackage {bbding}
\usepackage{makecell}
\usepackage{verbatim}
\usepackage{multirow}

\usepackage[frozencache,cachedir=minted-cache]{minted}
\usepackage{bm}
\usepackage[accsupp]{axessibility}  % Improves PDF readability for those with disabilities.

\usepackage[pagebackref,breaklinks,colorlinks]{hyperref}
\usepackage{orcidlink}

\newcommand{\topcaption}{%
    \setlength{\abovecaptionskip}{2pt}%
    \setlength{\belowcaptionskip}{0pt}%
\caption}

\makeatletter
\def\thanks#1{\protected@xdef\@thanks{\@thanks
        \protect\footnotetext{#1}}}
\makeatother

\begin{document}

\title{AddressCLIP: Empowering Vision-Language Models for City-wide Image Address Localization}

\titlerunning{AddressCLIP}

\author{Shixiong Xu\inst{1,3}$^{*}$$^{\dag}$\thanks{$^\dag$ This work was done when Shixiong Xu was an intern at Alibaba Cloud.} \and
Chenghao Zhang\inst{2}$^{ *}$\thanks{$^*$ Equal contributions} \and
Lubin Fan\inst{2}$^{ \ddag}$\thanks{$^\ddag$ Corresponding authors} \and
Gaofeng Meng\inst{1,3,4}$^{ \ddag}$ \and
Shiming Xiang\inst{1,3} \and
Jieping Ye\inst{2}}

\authorrunning{S. Xu et al.}

\institute{MAIS, Institute of Automation, Chinese Academy of Sciences \and
Alibaba Cloud \and
School of Artificial Intelligence, University of Chinese Academy of Sciences \and
CAIR, HK Institute of Science \& Innovation, Chinese Academy of Sciences}

\maketitle

\begin{abstract}
In this study, we introduce a new problem raised by social media and photojournalism, named \emph{Image Address Localization} (IAL), which aims to predict the readable textual address where an image was taken. 
Existing two-stage approaches involve predicting geographical coordinates and converting them into human-readable addresses, which can lead to ambiguity and be resource-intensive. In contrast, we propose an end-to-end framework named \emph{AddressCLIP} to solve the problem with more semantics, consisting of two key ingredients: i) image-text alignment to align images with addresses and scene captions by contrastive learning, and ii) image-geography matching to constrain image features with the spatial distance in terms of manifold learning. Additionally, we have built three datasets from Pittsburgh and San Francisco on different scales specifically for the IAL problem. Experiments demonstrate that our approach achieves compelling performance on the proposed datasets and outperforms representative transfer learning methods for vision-language models. Furthermore, extensive ablations and visualizations exhibit the effectiveness of the proposed method.
The datasets and source code are available at \url{https://github.com/xsx1001/AddressCLIP}.

\end{abstract}

\section{Introduction}
\label{sec:intro}
Users on social media platforms such as Facebook and Instagram often tag their pictures with textual addresses to connect with local communities, raising the demand for predicting the descriptive address information of the place where an image was taken. This has various practical applications, for instance, businesses and travel platforms can use addresses of images to provide recommendations or organize location-specific content. Additionally, photojournalism can rapidly verify the authenticity of the event with the image's address.

To predict an image's address, one reasonable approach 
involves leveraging image geo-localization technology to predict GPS coordinates (\ie, latitude and longitude) from an image~\cite{wilson2023image}, followed by the reverse Geocoding to query for a readable address. Image geo-localization, also known as visual place recognition, is commonly treated as an image retrieval problem where a database of geo-tagged images serves as a matching reference for the query image. Previous retrieval-based methods~\cite{NetVLAD,patch-netvlad,AnyLoc,MixVPR,CosPlace} have shown remarkable performance. However, in practice, the creation of pre-collected geo-tagged databases requires significant labor and storage resources, while GPS coordinates lack readability and semantics. In addition, the conversion from GPS to readable addresses often presents ambiguities, and the \emph{Image-GPS-Address} pipeline is not end-to-end.

\begin{figure*}[t]
\setlength{\abovecaptionskip}{-0.1cm}
\begin{center}
\includegraphics[width=1.0\textwidth]{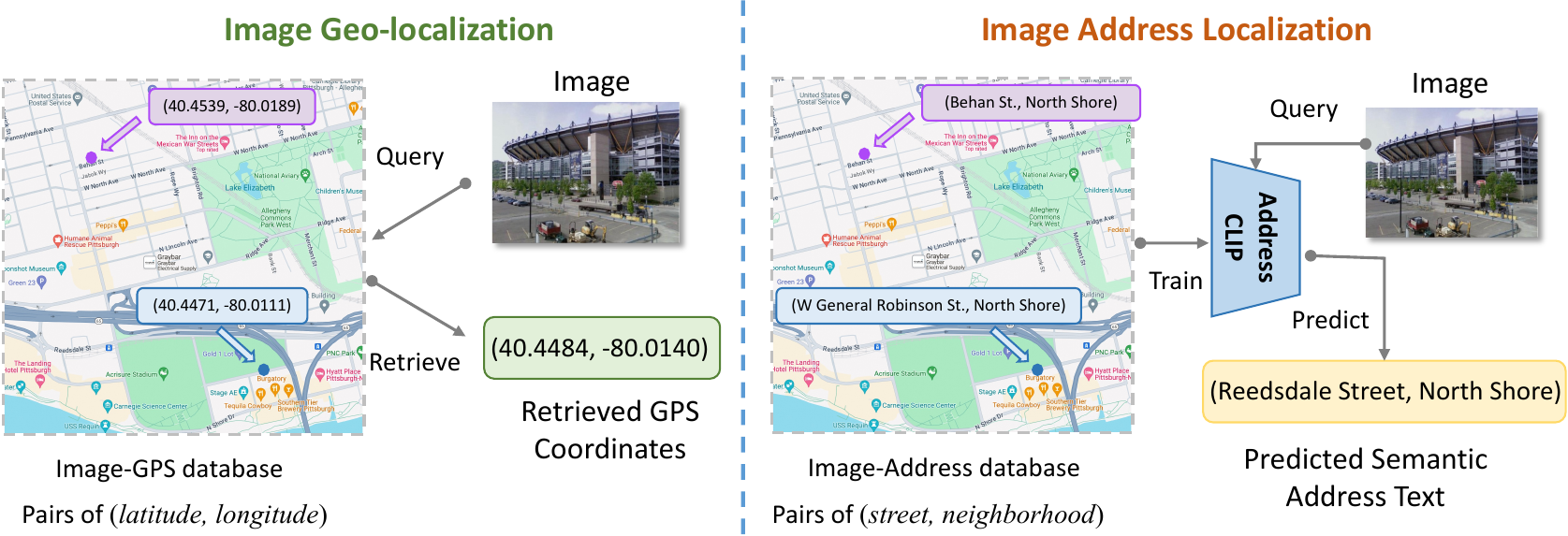}
\end{center}
\caption{Comparison of image-based geo-localization and address localization tasks. The objective of the proposed task is to predict the semantic text address of a given image instead of a digital GPS coordinate without the need for a retrieval gallery.}
\label{fig:intro_com}
\end{figure*}

To alleviate the above issues, in this study, we propose to perform \emph{Image Address Localization} (IAL) where a model is tasked to predict the readable textual address where a given image was taken. We design a \emph{semantic address partition} strategy to perform fine-grained partitioning of city-wide addresses, conforming to the way humans describe address information. By doing this, we are able to train models in an end-to-end manner, and during inference, there is no need to construct a retrieval database which greatly reduces the storage and retrieval burden. Furthermore, the model's output addresses align more closely with human description habits, which provides a bridge for subsequent city-wide scene understanding and point-of-interest recommendation. Fig.~\ref{fig:intro_com} shows the comparison of image geo-localization and image address localization tasks, where the latter focuses on predicting human-readable textual address information.

In this study, we propose an end-to-end framework, AddressCLIP, based on the visual-language model CLIP~\cite{CLIP}, aiming to learn an alignment of images and addresses. Our approach leverages two key ingredients: \emph{image-text alignment} and \emph{image-geography matching}.
Firstly, we introduce additional scene captions as a supplement to address text thus facilitating the alignment of images and textual addresses by contrastive learning. Secondly, we propose an image-geography matching mechanism to bring features of geographically proximate images closer while separating features of images that are far apart geographically.

To support the image address localization task, we constructed three IAL datasets of different sizes based on the Pitts-250k~\cite{NetVLAD} and SF-XL datasets~\cite{CosPlace}: Pitts-IAL (234K), SF-IAL-Base (184K), and SF- IAL-Large (1.96M). In contrast to the original datasets, each image in our dataset is accompanied by not only its geographical coordinate but also the administrative address. Specifically, we utilized the reverse Geocoding API of Google Maps to retrieve administrative addresses for a portion of the images and obtain addresses for the remaining images through nearest-neighbor interpolation of the geographical coordinates.

We evaluate the proposed AddressCLIP framework on the introduced datasets. Our proposed method achieves a Top-1 address localization accuracy of over 80\% across three IAL datasets, most notably reaching a performance of 85.92\% on the largest dataset, SF-IAL-Large. Compared with challenging baselines~\cite{zhou2022learning,zhou2022conditional,khattak2023maple} that transfer CLIP to the downstream IAL task, our AddressCLIP achieves improvements of 3\% to 6\% on the proposed datasets. In addition, the qualitative results demonstrate good alignment between images and textual address queries in geographical space. Finally, we discuss the superiority of the proposed method over the two-stage "Image-GPS-Address" approach and explore the application prospects of multimodal large language models in the IAL task.

Our contributions are summarized as follows:
\begin{itemize}
\item We formulate the image address localization problem and introduce the AddressCLIP framework for this problem by utilizing the alignment between the image and address text.
\item Two key ingredients are designed for better alignment of the image and address, \ie, image-caption alignment and image-geography matching, which are mutually beneficial.
\item We introduce three datasets named Pitts-IAL, SF-IAL-Base, and SF-IAL-Large to facilitate the study of the image address localization problem.
\item Experiments demonstrate that our method achieves
compelling performance on the proposed IAL datasets. Extensive ablations, visualizations, and analyses are provided to show the effectiveness of the proposed method.
\end{itemize}

\begin{figure*}[t]
\setlength{\abovecaptionskip}{-0.1cm}
\begin{center}
\includegraphics[width=1.0\textwidth]{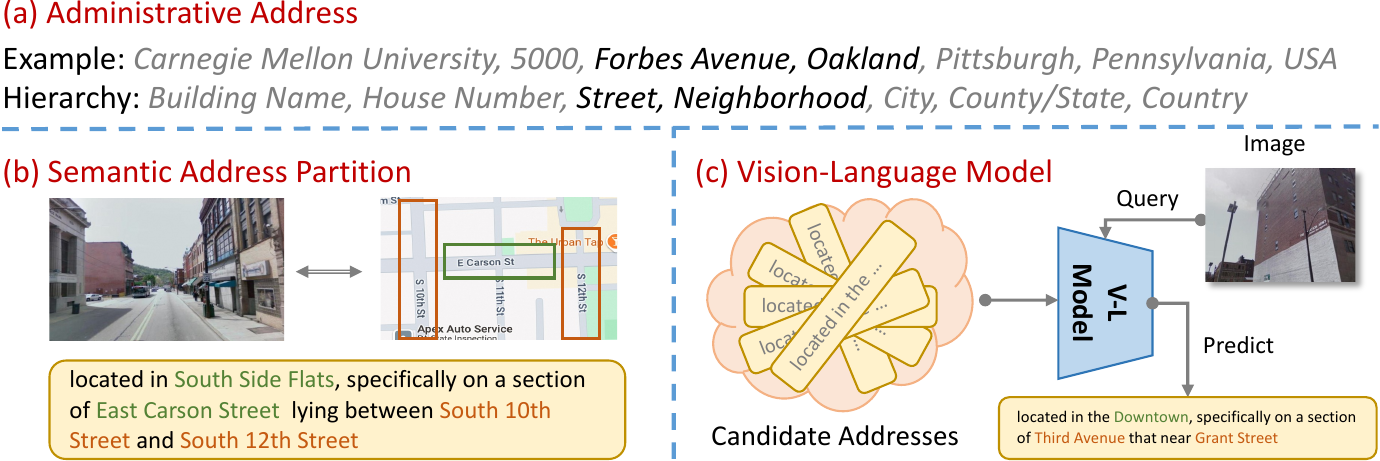}
\end{center}
\caption{The problem statement of the image address localization task consists of (a) examples of administrative address and hierarchy, (b) semantic address partition strategy, and (c) address predicting using visual-language models.}
\label{fig:problem}
\end{figure*}

\section{Related Work}

\noindent\textbf{\emph{Image Geo-localization.}}
Image geo-localization, or visual place recognition, is usually formulated as an image retrieval problem on the city scale, which needs to collect a geo-tagged database of pre-computed embeddings of either local or global features~\cite{local2004,Surf,global2006,vpr2004,vpr2008,VLAD,vpr2007,pitts}. In recent years, deep learning models~\cite{VGG,ResNet,ViT} have been proven to perform remarkably in image feature extraction, complemented with an aggregation or pooling layer~\cite{NetVLAD,CRN,improve_vlad1,R-MAC,GEM,improve_vlad2,APANet,SFRS,MixVPR}. Recent methods achieve impressive retrieval performance by performing an additional re-ranking phase~\cite{patch-netvlad,TransVPR,r2former}, adopting powerful pre-trained backbones~\cite{dinov2} to extract image features~\cite{AnyLoc,sota}, or training on large-scale place recognition datasets~\cite{CosPlace,MixVPR,GSV-cities,sota}. Different from retrieval-based methods, classification-based methods focus on planet-scale localization and split the earth into disjoint regions to classify~\cite{divide&classify,planet,cplanet,GeoLocator,hierarachies}. More recently, StreetCLIP~\cite{StreetCLIP} and GeoCLIP~\cite{GeoCLIP} both utilize the vision-language model CLIP~\cite{CLIP} with region description or GPS information for better generalizability. 
Going beyond image geo-localization, we propose to perform image address localization to obtain readable textual addresses rather than digital coordinates without a retrieval gallery. This not only enables models to directly output human-understandable semantic addresses for a given image but also paves the way for more complex geographical human-computer interactions in the future.

\noindent\textbf{\emph{Transfer Learning in Vision-Language Models.}}
The integration of language supervision with visual data is garnering significant interest, with the primary aim being to align images and texts and learn a shared embedding space. As outlined in~\cite{zhou2022learning}, the advancements in vision-language models can largely be attributed to three key developments: Transformers~\cite{vaswani2017attention}, contrastive representation learning~\cite{chen2020simple,he2020momentum}, and expansive web-scale training datasets~\cite{jia2021scaling,radford2021learning}. One notable example is CLIP~\cite{CLIP}, which employs two encoder networks trained via contrastive loss to align image-text pairs, thus enabling impressive zero-shot performance. Adapting CLIP to downstream tasks typically involves either full fine-tuning or linear probing~\cite{gao2023clip}. Recently, prompt learning offers an alternative by introducing a small number of trainable prompt tokens at the input. Learnable prompts can be applied to the language branch~\cite{zhou2022learning}, image instances~\cite{zhou2022conditional}, or both forming a multi-modal prompt~\cite{khattak2023maple}. 
Complete fine-tuning enables CLIP to fully adapt to the data distribution of downstream tasks, while prompt learning enhances CLIP's zero-shot learning capabilities. 
Due to the domain gap between the IAL task and the pre-training tasks, our proposed AddressCLIP 
adopts carefully designed image-caption alignment and image-geography matching to transfer CLIP toward the address localization task, which is superior to the direct complete fine-tuning way.

\section{Problem Statement}
\label{sec:problem}
In this study, we focus on the city-wide image address localization problem. The administrative address hierarchy around the world varies widely depending on the history, geography, culture, and political systems of each country. 
Taking the United States as an example, we provide a specific illustration of an administrative address and its corresponding hierarchy in Fig.~\ref{fig:problem} (a). Since images in one dataset belong to the same city, our study distinguishes image addresses on \emph{neighborhood} and \emph{street} levels.

The straightforward division mentioned above introduces two challenges in practical city-wide scenarios. Firstly, variable street lengths can result in coarsely localized addresses,  
particularly for highways that extend for kilometers,
creating a pronounced long-tail distribution issue and diverse inner-address visual features that hinder precise localization during inference. Secondly, address ambiguity arises at street intersections, where images could be equally attributed to intersecting streets, thus lacking a clear and singular textual supervision signal.
To address these concerns, we introduce a \emph{semantic address partition} strategy for a more granular segmentation of streets, as shown in Fig.~\ref{fig:problem} (b). By segmenting streets at intersections, we achieve a balance in street lengths, which refines the address localization scope and eliminates the intersection ambiguity, aligning more closely with the way humans typically describe locations. In this way, the textual representation of addresses consists of the main street name (marked \textcolor{green}{green}) and the name of one or two streets that intersect it (marked \textcolor{brown}{brown}).

Formally, the \emph{Image Address Localization} problem is defined as follows: given a training dataset $D_{train} = \{(I_i, A_i)\}_{i=1}^M$ containing pairs of image $I_i$ and address $A_i$, our objective is to train a vision-language model $\mathcal{H}_{\theta}$ and use it to predict the address of query images, $A_k^Q = \mathcal{H}_{\theta}(I_k^Q), \forall k \in [1..K]$ where $I_k^Q \in D_{test}$.
The images in the query set $I^Q$ can belong to any candidate address in the same city as the images in the training set. Fig.~\ref{fig:problem} (c) shows a schematic diagram of predicting the readable textual address for a given query image.

\begin{figure*}[t]
\setlength{\abovecaptionskip}{-0.1cm}
\begin{center}
\includegraphics[width=1.0\textwidth]{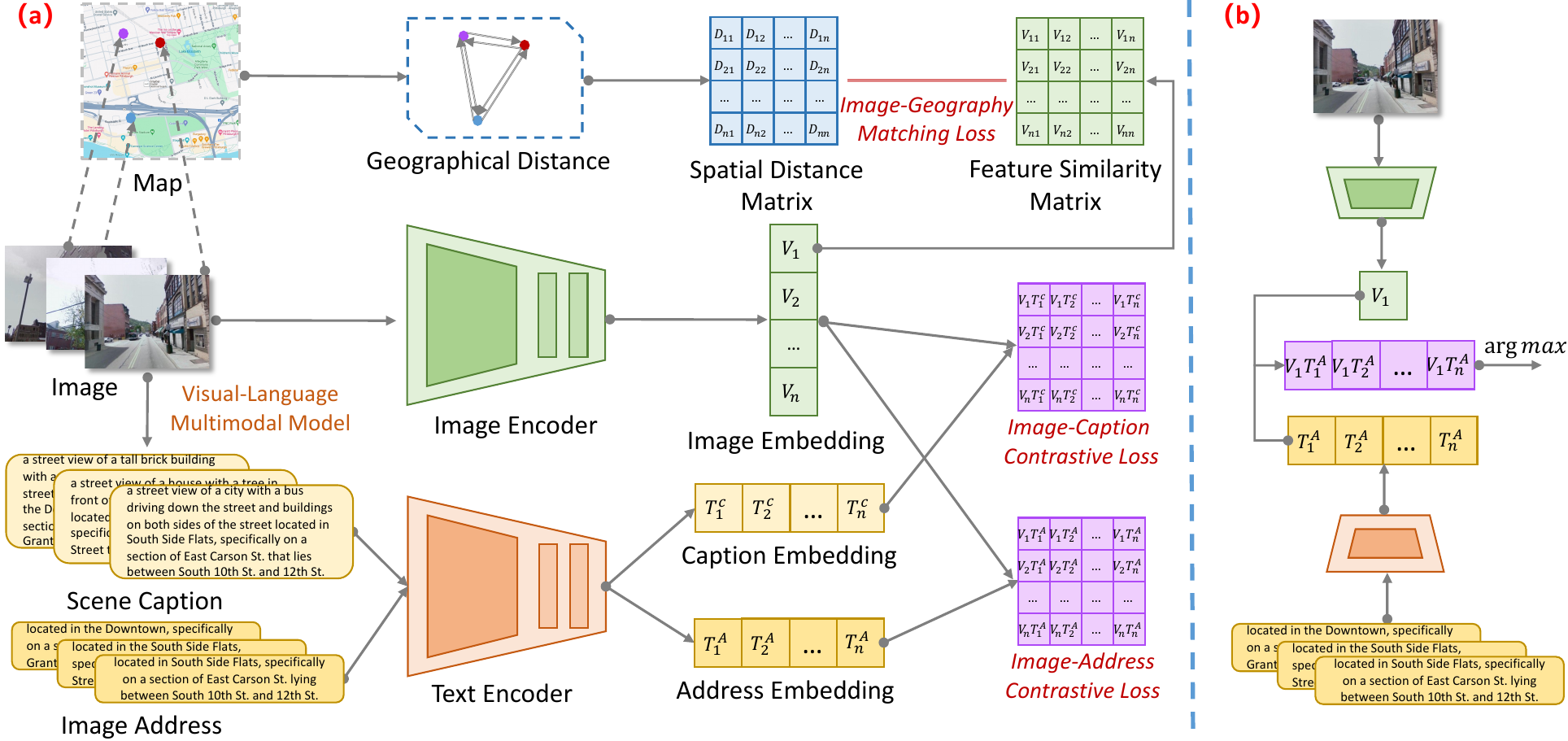}
\end{center}
\caption{Overview of the proposed AddressCLIP framework. (a) During training, the alignment of image and address is learned by the image-address contrastive loss, image-caption contrastive loss, and image-geography matching loss. (b) At inference, the address with the highest similarity to the query image’s embedding is chosen.}
\label{fig:overview}
\end{figure*}

\section{AddressCLIP}
\label{sec:method}
\subsection{Framework Overview}
We formulate the IAL problem as a vision-text alignment problem between the image and address pairs. Fig.~\ref{fig:overview} depicts the framework of our method. 
During training, the embeddings of the image and the address are extracted by the image encoder and text encoder, respectively, and are then aligned through image-address contrastive learning. An additive scene caption is introduced as a supplement to the address to enrich the plain text information. The scene caption shares the same text encoder with the image address, and the resulting caption embedding and image embedding are combined for image-caption contrastive learning. Furthermore, we adopt the geographical position information as a guide to increase the similarities between geographically close image features while increasing the differences between geographically distant image features.
The image-geography matching is learned between geospatial distance and image feature similarity. 
During inference, the address with the highest similarity to the query image’s embedding indicates the most probable address.

\subsection{Image-Text Alignment}
It is reasonable to use address information directly as textual prompts for image-address alignment learning. However, the address text is simple and limited. It cannot provide context about environments, landmarks, or other entities, which are crucial for precise address localization. To alleviate the issues, we incorporate additional descriptive captions that capture the nuances of the visual scene, thereby endowing the model with a deeper understanding of the contextual elements that are often missing in the bare address labels. This mechanism enables more accurate and context-aware predictions by effectively bridging the gap between visual perception and textual representation.

Scene description can be generated through manual annotation, which, although accurate, is costly and not easily scalable to large datasets. Benefiting from the advancements in vision-language models, we utilize pre-trained vision-language models~\cite{li2022blip} to generate linguistic captions corresponding to image scenes.
The lower left corner of Fig.~\ref{fig:overview}(a) shows some illustrative examples, where the descriptions can include context like the presence of specific buildings or unique street signs, which is relevant for distinguishing between visually similar but geographically distant locations. This also aligns the model's learning process with how humans typically communicate location information. For detailed analyses of scene captions, refer to the appendix.

Formally, define image features extracted from the image encoder $\mathcal{V}(\cdot)$ as $V_i = \mathcal{V}(I_i), \forall i \in [1,...,N]$. The text encoder $\mathcal{V}(\cdot)$ extracts address features $T^A_i = \mathcal{T}(A_i)$ and caption features $T^C_i = \mathcal{T}(C_i + A_i)$, where the scene caption $C_i$ is obtained by a vision-language model. We experimentally observe that appending address information to the scene caption is more conducive to address localization, which is discussed in detail in Sec.~\ref{sec:ablation}. Note that the additive scene caption is only used for training. The alignment of images and addresses is learned via \emph{image-address contrastive loss} and \emph{image-caption contrastive loss}.

For a batch of size $N$ comprising image-text pairs, the image-address contrastive loss can be written as:
\begin{equation}
\small
\mathcal{L}_{address} = - \frac{1}{2N} \sum_{i=1}^{N} \left[ \log \frac{\exp(V_i \cdot T^A_i / \tau)}{\sum_{j=1}^{N} \exp(V_i \cdot T^A_j / \tau)} + \log \frac{\exp(T^A_i \cdot V_i / \tau)}{\sum_{k=1}^{N} \exp(T^A_i \cdot V_k / \tau)} \right],
\end{equation}
where $\tau$ is the temperature parameter. Similarly, the image-caption contrastive loss is formulated as:
\begin{equation}
\small
\mathcal{L}_{caption} = - \frac{1}{2N} \sum_{i=1}^{N} \left[ \log \frac{\exp(V_i \cdot T^C_i / \tau)}{\sum_{j=1}^{N} \exp(V_i \cdot T^C_j / \tau)} + \log \frac{\exp(T^C_i \cdot V_i / \tau)}{\sum_{k=1}^{N} \exp(T^C_i \cdot V_k / \tau)} \right].
\end{equation}

\subsection{Image-Geography Matching}
In general, address text in city-wide scenarios may be geographically far away but highly similar, or geographically close but significantly different. This makes image-address alignment learning difficult to optimize with address text alone. In contrast, the geographic coordinates of images (e.g., UTM coordinates) differ significantly, showcasing clear distinctions and discriminative properties. From the perspective of manifold learning, image embedding represents a low-dimensional representation of images in the feature space, and its distribution should be consistent with the geographic coordinates of the images. Our goal is to ensure that geographically proximate images exhibit closely in the feature space, while geographically distant images reflect more within the feature space. Visualization results and analysis are elaborated in Sec.~\ref{sec:visulization}.

Inspired by the above motivation, we propose an \emph{image-geography matching loss} to constrain image features according to the spatial distances of geographic coordinates. Specifically, denote $U_i:\mathbb{UTM}_{east} \times \mathbb{UTM}_{north}, \forall i \in [1,...,N]$ the set of geographic coordinates corresponding to all images within a batch of size $N$. We can calculate each element of the spatial distance matrix $D^U$ in the geographic space as follows:
\begin{equation}
D^U_{ij} = ||\hat{U}_i - \hat{U}_j||_1,\, s.t., \, \hat{U}_i = \frac{U_i - \min(U_i)}{\max(U_i) - \min(U_i)},
\end{equation}
where Manhattan distance and min-max normalization are adopted. Correspondingly, each element of the feature similarity matrix $D^V$ in the image embedding space is calculated as:
\begin{equation}
D^V_{ij} = \frac{V_i \cdot V_j}{||V_i|| \cdot ||V_j||}.
\end{equation}
Consequently, the image-geography matching loss takes the image feature similarity matrix $D^V$ as input and the geographic spatial distance matrix $D^U$ as the target to perform gradient back-propagation, \ie,
\begin{equation}
\mathcal{L}_{geography} = \frac{1}{N^2} \sum_{i=1}^{N} \sum_{j=1}^{N} (D^V_{ij} - D^U_{ij})^2.
\end{equation}

\begin{table}[tb]
  \topcaption{Detailed information of the proposed Image Address Localization datasets.}
  \label{tab:dataset}
  \centering
  \scriptsize
  \tabcolsep=0.12cm
  \begin{tabular}{lcccccccc}
    \toprule
    Dataset & Year & \makecell[c]{Dataset \\ size} & \makecell[c]{\# \\ train/val} & \makecell[c]{\# \\ test} &  \makecell[c]{Query \\ type} & \makecell[c]{Image \\ size} & GPS & Address\\
    \midrule
    Pitts-250K~\cite{NetVLAD} & 2016 & 9.4GB & 250K & 24K & panorama & 480$\times$640 & \ding{52} & \ding{56} \\
    SF-XL~\cite{CosPlace}  & 2022 & 1TB & 41.2M & 1K/0.6K & phone & 512$\times$512 & \ding{52} & \ding{56} \\
    \midrule
    Pitts-IAL & 2024 & 6.7GB & 234K & 19K & panorama & 480$\times$640 & \ding{52} & \ding{52} \\
    SF-IAL-Base  & 2024 & 6.8GB & 184K & 21K & panorama & 512$\times$512 & \ding{52} & \ding{52} \\
    SF-IAL-Large & 2024 & 121GB & 1.96M & 280K & panorama & 512$\times$512 & \ding{52} & \ding{52} \\
  \bottomrule
  \end{tabular}
\end{table}

\begin{figure*}[t]
\setlength{\abovecaptionskip}{-0.1cm}
\begin{center}
\includegraphics[width=0.98\textwidth]{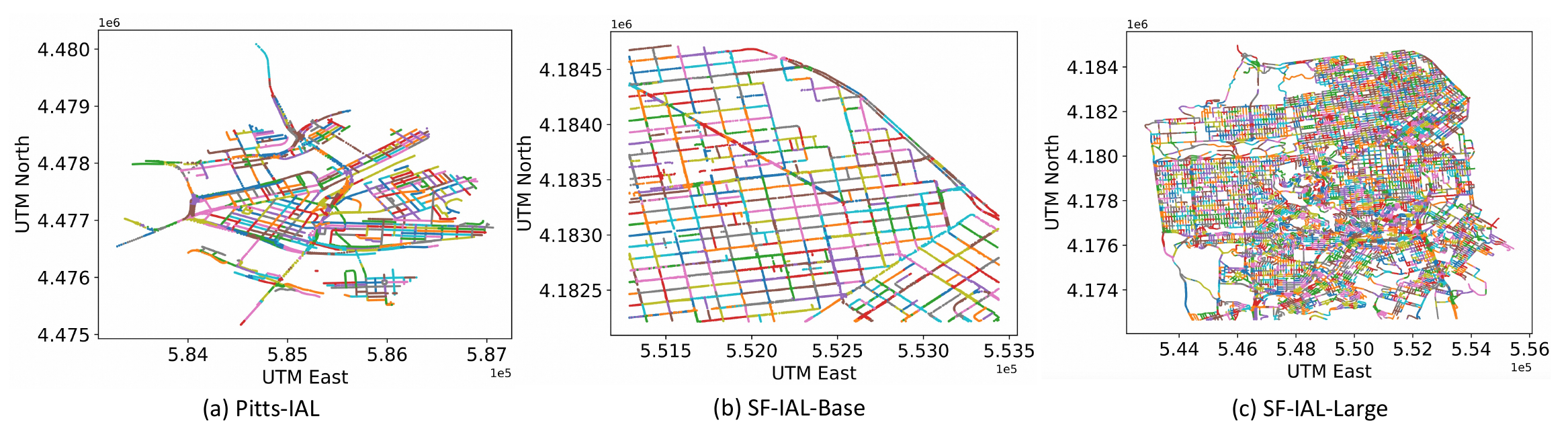}
\end{center}
\caption{Visualizations of the introduced datasets. Distinct semantic street partitions are displayed using varying colors.}
\label{fig:datasets}
\end{figure*}

\subsection{Objective Function}
We train the proposed AddressCLIP using both image-text contrastive loss and image-geography matching loss in an end-to-end manner. The total objective function is as follows:
\begin{equation}
\mathcal{L}_{total} = \alpha \mathcal{L}_{address} + \beta \mathcal{L}_{caption} + \gamma \mathcal{L}_{geography},
\end{equation}
where $\alpha$, $\beta$, and $\gamma$ are weight parameters.

\section{Image Address Localization Datasets}
\label{sec:dataset}
Existing datasets~\cite{CosPlace, pitts,GSV-cities} for image geo-localization only contain the GPS coordinates of where the image was taken. Meanwhile, the text in popular image-text datasets like LAION-5B~\cite{LAION-5B} mainly describes the semantic content of the corresponding image instead of the geographical information. To support the study of the IAL problem, we introduce three IAL datasets named Pitts-IAL, SF-IAL-Base, and SF-IAL-Large derived from Pitts-250k~\cite{pitts} and SF-XL~\cite{CosPlace}, respectively. We describe the details of how these datasets were built below.

\subsection{Address Annotation}
We look up the administrative address according to the GPS coordinates attached to images by utilizing the Reverse Geocoding API of Google Maps. The API returns a list of addresses ordered by their match degree with the GPS coordinate, \eg $[A^{(1)}, A^{(2)}, \cdots, A^{(R)}]$. However, simply selecting $A^{(1)}$ as the address annotation is often imprecise since the API might match the GPS coordinates of a building's center and return the building's address. Additionally, when a building is located at an intersection of cross streets, the API might return ambiguous addresses. To alleviate the issue, we first exclude address information matched to buildings (labeled as "ROOFTOP" location type in the API). Then, we choose the most frequently occurring address among the remaining addresses as the definitive address and ensure its accuracy by random manual verification and correction. Finally, we adopt the introduced semantic address partition strategy for fine-grained partitioning as the final address annotation.

\subsection{Statistics and Visualization}
We provide a comprehensive comparison between the proposed IAL datasets in~\cref{tab:dataset} and visualize their street distributions in~\cref{fig:datasets}.
Specifically, \textbf{Pitts-IAL} is constructed using the training set of the original Pitts-250K~\cite{pitts} dataset where 10,586 locations are annotated with 24 images from different views for each location. These image-address pairs are divided into a training, database, and {query set randomly using a ratio of 7:2:1 according to locations. Due to the sparseness of the Pitts-250K, the queries are filtered to ensure their address can be covered by the training set and database. \textbf{SF-IAL} is constructed from the SF-XL~\cite{CosPlace} dataset and is divided into two versions according to the size of the coverage area, namely SF-IAL-Base, and SF-IAL-Large. SF-IAL-Base covers the top-right corner of San Francisco with 17,067 locations, each with 12 images from different views, which is of comparable size to Pitts-IAL. SF-IAL-Large covers the entire San Francisco with 233,820 locations. 
The image-address pairs in both versions are also divided into a training, database, and query set randomly using a ratio of 7:2:1. The datasets introduced have been released to the community for research at \url{https://github.com/xsx1001/AddressCLIP}.

\section{Experiments}
\label{sec:exp}

\begin{table}[!htbp]
\begin{center}
\topcaption{Evaluation results of address localization on the Pitts-IAL, SF-IAL-Base, and SF-IAL-Large datasets.}
\label{tab:main_result}
% \tiny
\scriptsize
%\tablestyle{2.2mm} {1.1}{
\setlength{\tabcolsep}{0.15mm}{
\begin{tabular}{ c  | c c c c | c c c c | c c c c}
\toprule
\multirow{2}{*}{Method}  & \multicolumn{4}{c|}{Pitts-IAL} & \multicolumn{4}{c|}{SF-IAL-Base} & \multicolumn{4}{c}{SF-IAL-Large} \\
& SSA-1 &  SSA-5 & SA-1 &  SA-5 & SSA-1 & SSA-5 & SA-1 & SA-5 & SSA-1 & SSA-5 & SA-1 & SA-5 \\
\midrule
Zero-shot CLIP & 0.85 & 3.69 & 1.28 & 5.64 & 1.25 & 5.30 & 2.80 & 9.06 & 0.26 & 0.97 & 0.50 & 2.85\\
CLIP + address & 77.66 & 93.28 & 80.86 & 94.17 & 83.66 & 96.32 & 85.76 & 96.85 & 81.84 & 95.38 & 84.56 & 95.79\\
\hline
CLIP + CoOp~\cite{zhou2022learning} & 67.91 & 86.60 & 71.19 & 88.18 & 77.77 & 94.05 & 79.90 & 94.91 & 74.84 & 92.38 & 78.23 & 93.79\\
CLIP + CoCoOp~\cite{zhou2022conditional} & 69.04 & 88.34 & 73.28 & 89.78 & 79.19 & 95.27 & 81.15 & 96.32 & 76.92 & 93.58 & 79.85 & 94.04\\
CLIP + MaPLe~\cite{khattak2023maple} & 72.98 & 91.85 & 76.04 & 92.27 & 81.46 & 96.98 & 83.69 & 97.77 & 79.63 & 94.47 & 82.34 & 95.96\\
\midrule
\textbf{AddressCLIP (Ours)} & \textbf{80.39} & \textbf{96.27} & \textbf{82.62} & \textbf{96.74} & \textbf{86.32} & \textbf{99.09} & \textbf{87.44} & \textbf{99.23} & \textbf{85.92} & \textbf{97.28} & \textbf{88.10} & \textbf{98.33}\\
\bottomrule
\end{tabular}}
\end{center}
\end{table}

\subsection{Experimental Setup}
\label{sec:exp_setup}
\noindent\textbf{\emph{Implementation Details.}}
Our AddressCLIP is implemented with PyTorch based on the pre-trained CLIP from OpenAI~\cite{CLIP} with no additional parameters. All the images are resized to 224$\times$224 and normalized to fit the input of CLIP. Unless otherwise stated, the ViT/B-16 version of CLIP is used for experiments. We adopt the vision-language model BLIP~\cite{li2022blip} to generate additive scene captions. More training details are given in the appendix.

\noindent\textbf{\emph{Metrics.}} 
It is straightforward to measure the address localization performance by calculating the accuracy of the predicted address, like standard Top-1 and Top-5 accuracy. Considering the varying precise requirements for the returned addresses in different scenarios, we design two metrics specifically for evaluating the address localization performance, \ie, \textit{Street-level Accuracy (SA)} and \textit{Sub-Street-level Accuracy (SSA)}. Formally, for a given query image, the output of the model could be denoted by $A_p = [S^m, S^c, S^n]$, where $S^m$ is the main street, $S^c$ is the set of streets that intersect with $S^m$, and $S^n$ is the neighborhood. The groundtruth address is denoted by $A_{gt} = [S^m_{gt}, S^c_{gt}, S^n_{gt}]$. If $S^m=S^m_{gt}$ and $S^n=S^n_{gt}$, the prediction is correct in street-level. It is correct in the sub-street level only when $A_p=A_{gt}$ is satisfied. Both Top-1 and Top-5 accuracy are reported as SA-1, SA-5, SSA-1, and SSA-5.

\subsection{Main Results}
\label{sec:exp_main}
\noindent\textbf{\emph{Baselines.}} 
We compare our method with zero-shot CLIP and a fine-tuned CLIP model with naive address prompts. Image address localization can be considered a downstream visual-language task thus prompt learning approaches can be used to transfer the pre-trained CLIP to address localization. We also compare with several representative prompt learning methods for visual-language models, \ie, CoOp~\cite{zhou2022learning}, CoCoOp~\cite{zhou2022conditional}, and MaPLe~\cite{khattak2023maple}.

\noindent\textbf{\emph{Comparisons.}}
Tab.~\ref{tab:main_result} shows the comparison results with the above baselines on the introduced Pitts-IAL, SF-IAL-Base, and SF-IAL-Large datasets. It is clear that our method achieves remarkable performance on the three datasets across various metrics. The zero-shot CLIP model exhibits poor performance due to the lack of explicit address information in the image-text pairs during pre-training. After fine-tuning CLIP with address, the address localization accuracy improves significantly on all three datasets, forming a strong baseline.

\begin{table}[!htbp]
\begin{center}
\topcaption{Ablation study of key components on the proposed datasets.}
\label{tab:key_components}
\scriptsize
\setlength{\tabcolsep}{1.0mm}{
\begin{tabular}{ c  c  c| c  c c c |cccc }
\toprule
\multirow{2}{*}{$\mathcal{L}_{address}$} & \multirow{2}{*}{$\mathcal{L}_{caption}$} & \multirow{2}{*}{$\mathcal{L}_{geography}$}  & \multicolumn{4}{c|}{Pitts-IAL} & \multicolumn{4}{c}{SF-IAL-Base}  \\
& & & SSA-1 &  SSA-5 & SA-1 &  SA-5 & SSA-1 & SSA-5 & SA-1 & SA-5 \\
\midrule
\ding{52} & &  & 77.66 & 93.28 & 80.86 & 94.17 & 83.66 & 96.32 & 85.76 & 96.85  \\
 & \ding{52} & & 69.27 & 87.23 & 71.39 & 88.92 & 75.85 & 89.21 & 77.24 & 91.46 \\
\ding{52} & \ding{52} &  & 79.20 & 94.15 & 81.26 & 94.64 & 84.86 & 97.46 &86.03 &98.04\\
\ding{52} & & \ding{52} & 79.27 & 95.15 & 81.45 & 95.61 &85.54 & 98.98 & 86.64 & 98.15\\
\ding{52} & \ding{52} & \ding{52} & \textbf{80.39} & \textbf{96.27} & \textbf{82.62} & \textbf{96.74} & \textbf{86.32} & \textbf{99.09} & \textbf{87.44} & \textbf{99.23}\\
\bottomrule
\end{tabular}}
\end{center}
\end{table}

\begin{table}[!htbp]
\begin{center}
\scriptsize
\topcaption{Performance of different encoder training strategies on the proposed datasets. \ding{56} refers to freezing the weight, and \ding{52} refers to unfreezing the weight.}
\label{tab:encoder}
\setlength{\tabcolsep}{2.2mm}{
\begin{tabular}{ c c| c c c c |cccc}
\toprule
\multirow{2}{*}{Image} & \multirow{2}{*}{Text} & \multicolumn{4}{c|}{Pitts-IAL} & \multicolumn{4}{c}{SF-IAL-Base}  \\
& & SSA-1 &  SSA-5 & SA-1 &  SA-5 & SSA-1 & SSA-5 & SA-1 & SA-5 \\
\midrule
\ding{52} & \ding{56} & 77.77 & 89.20 & 80.28 & 90.48 &84.32 & 93.63 & 85.82 & 95.05 \\
\ding{56} & \ding{52} & 48.88 & 78.31 & 52.43 & 80.89 & 54.62 & 83.74 & 57.50 & 86.06\\
\ding{52} & \ding{52} & \textbf{80.39} & \textbf{96.27} & \textbf{82.62} & \textbf{96.74} & \textbf{86.32} & \textbf{99.09} & \textbf{87.44} & \textbf{99.23} \\
\bottomrule
\end{tabular}}
\end{center}
\end{table}

Benefiting from carefully designed image-text alignment and image-geography matching mechanisms, our AddressCLIP surpasses the representative visual-language prompt learning methods by 7.41\%, 4.86\%, and 6.29\% on Pitts-IAL, SF-IAL-Base, and SF-IAL-Large datasets respectively in terms of SSA-1. This indicates that general prompt learning methods that transfer pre-trained models to various downstream tasks are inferior to those specifically designed, especially when the domain of the downstream task (IAL) differs significantly from that of the pre-trained.
It is noteworthy that our method generally performs better on the SF-IAL-Base dataset than on the Pitts-IAL dataset due to more orderly streets and the greater density of street view image collection. Remarkably, our method achieves an address location accuracy of 85.92\% even on the more challenging SF-IAL-Large dataset, which covers an area $8 \times$ larger than the Pitts-IAL dataset.
Additionally, performance on the SA metric is typically higher than the SSA metric, suggesting that using sub-streets as the learning target can further enhance the localization capability for main streets.

\subsection{Ablation Study}
\label{sec:ablation}
\noindent\textbf{\emph{Effectiveness of Key Components.}}
We adopt the CLIP model fine-tuned with the image-address contrastive loss $\mathcal{L}_{address}$ as the ablation baseline to show the effectiveness of proposed image-caption alignment and image-geography matching. The evaluation results on the Pitts-IAL and SF-IAL-Base datasets are listed in Tab.~\ref{tab:key_components}. As can be seen, applying $\mathcal{L}_{caption}$ alone yields reasonable address localization accuracy, but it is far inferior to using $\mathcal{L}_{address}$ alone, suggesting that the independent role of address information better facilitates image-address alignment. Based on $\mathcal{L}_{address}$, adding $\mathcal{L}_{caption}$ increases SSA-1 by 1.54\% and 1.2\%, while adding $\mathcal{L}_{geography}$ alone increases by 1.61\% and 1.88\% on the two IAL datasets. This demonstrates that both the proposed mechanisms can facilitate image address alignment learning from their respective perspectives. Their combination can ultimately bring 2.73\% and 2.66\% improvement on SSA-1, indicating a mutually beneficial relationship between them. Consistent conclusions can be drawn on other indicators.

\begin{figure*}[t]
\setlength{\abovecaptionskip}{-0.2cm}
\begin{center}
\includegraphics[width=0.98\textwidth]{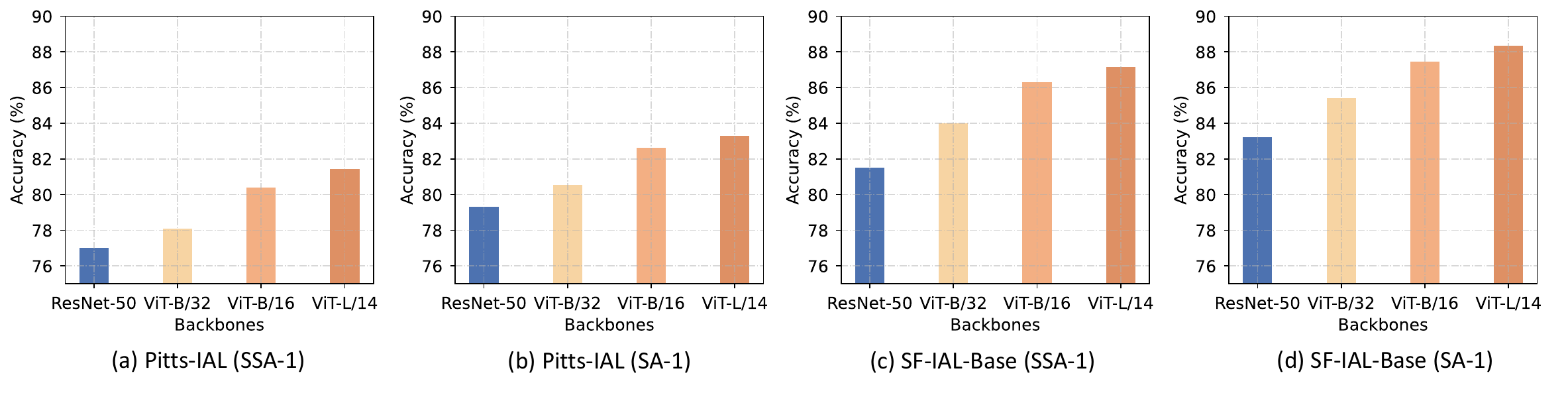}
\end{center}
\caption{Performance of different backbones on the proposed datasets.}
\label{fig:backbone}
\end{figure*}

\begin{figure*}[t]
\setlength{\abovecaptionskip}{-0.2cm}
\begin{center}
\includegraphics[width=0.98\textwidth]{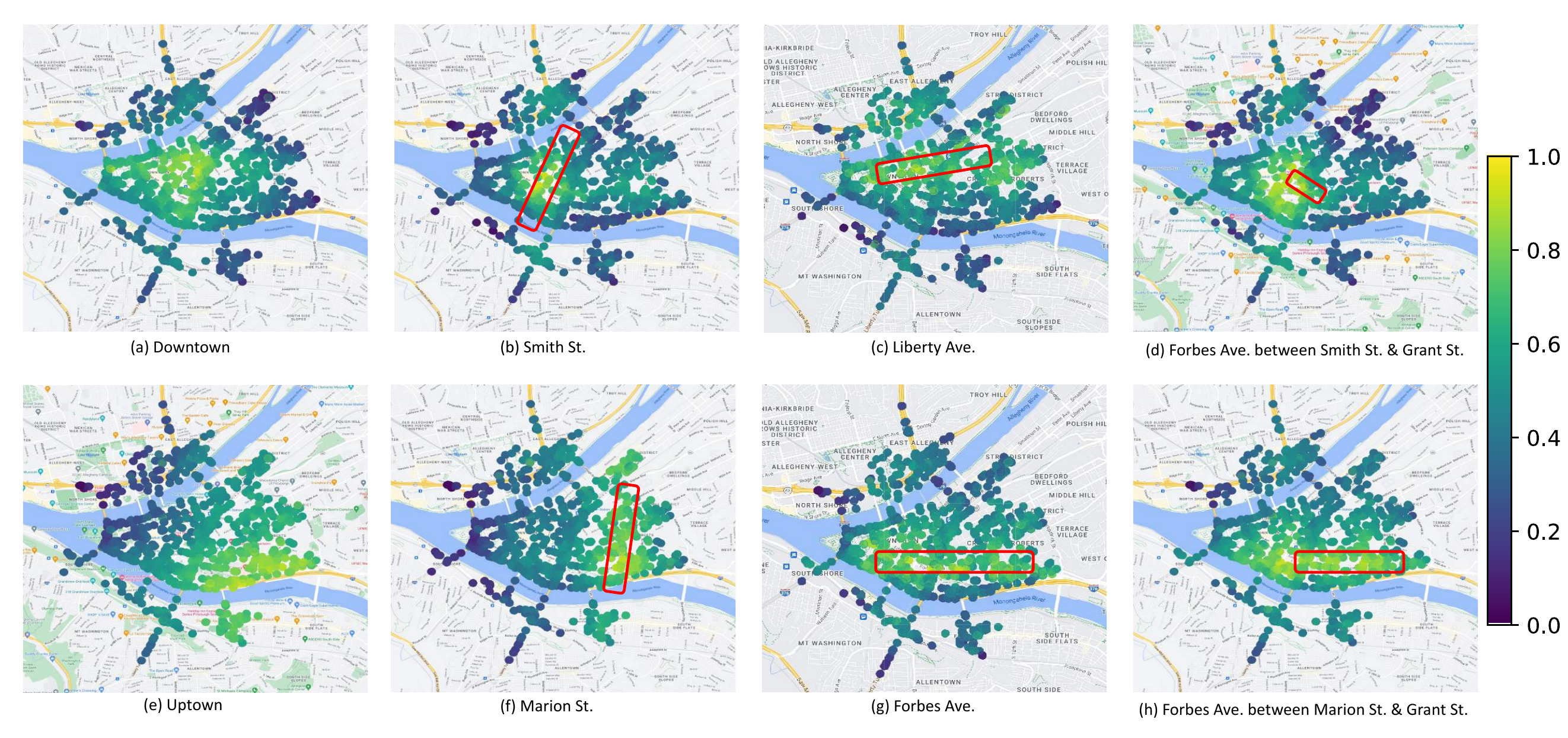}
\end{center}
\caption{Qualitative demonstration: Address localization with a given textual address query using AddressCLIP in Pittsburgh. The brighter the scatter point, the higher the similarity of the embedding between the image and the query address text. The red box represents the actual geographic range of the query street in the map.}
\label{fig:visualization}
\end{figure*}

\noindent\textbf{\emph{Encoder Training Strategy.}}
Typically, when adapting CLIP to downstream tasks, the impact of unfreezing the weights of the image and text encoders varies on the outcomes. Tab.~\ref{tab:encoder} shows the performance comparisons of different encoder freezing strategies on the Pitts-IAL and SF-IAL-Base datasets. It is observable that unfreezing only the image encoder brings much more performance gains (about 30\%) compared to unfreezing only the text encoder. This suggests that visual discrepancies are more prominent than textual ones when transferring CLIP to the task of address localization. The best performance is achieved when both the image encoder and text encoder are concurrently unfrozen. This is consistent with the intuitive notion that the textual address is significantly different from the natural category CLIP has been pre-trained on, necessitating the unfreezing of more weights for fine-tuning.

%\vspace{1pt}
\begin{table}[!htbp]
\begin{center}
\topcaption{Comparisons with retrieval-based image geo-localization methods in terms of storages and time overheads, without considering the API query time.}
\label{tab:time_storage_comparison}
\scriptsize
\setlength{\tabcolsep}{2.2mm}{
\begin{tabular}{  c | c  c  c  c   c }
\toprule
Methods & Storage & Inference & Retrieval & Reranking & Memory \\
\midrule
TransVPR~\cite{TransVPR} & 2.02 GB & 6.20 ms & 0.19 ms & 1757.70 ms & 61.12 GB \\
R2Former~\cite{r2former} & 2.10 GB & 8.81 ms & 0.19 ms & 202.37 ms & 12.64 GB \\
SALAD~\cite{SALAD} & 2.34 GB & \textbf{2.34 ms} & 0.19 ms & \textbf{0} & 1.69 GB \\
\midrule
\textbf{AddressCLIP} &\textbf{0.34 GB} & 3.46 ms & \textbf{0} & \textbf{0} & \textbf{0.64 MB}\\
\bottomrule
\end{tabular}}
\end{center}
\end{table}

\begin{table}[!htbp]
\begin{center}
\topcaption{Performance comparisons with retrieval-based image geo-localization methods using the reverse Geocoding API
on the Pitts-IAL dataset.}
\label{tab:retieval_with_api}
\scriptsize
\setlength{\tabcolsep}{0.5mm}{
\begin{tabular}{ c | c  c  c c c | c  }
\toprule
Methods & CosPlace\cite{CosPlace} & MixVPR\cite{MixVPR} & EigenPlaces\cite{EigenPlaces} & AnyLoc\cite{AnyLoc} & SALAD\cite{SALAD} & \textbf{AddressCLIP} \\
\midrule
SSA-1 & 73.04 & 74.52 & 73.88 & 74.83 & 75.17 & \textbf{77.01}\\
SSA-5 & 92.43 & 93.67 & 93.79 & 93.45 & 94.23 & \textbf{95.33}\\
\bottomrule
\end{tabular}}
\end{center}
\end{table}

\noindent\textbf{\emph{Different Backbones.}}
Fig.~\ref{fig:backbone} shows the performance using different backbones on the Pitts-IAL dataset. We adopt Transformer-based ViT-B/16, ViT-B/32, and ViT-L/14~\cite{ViT}, as well as ResNet-50~\cite{ResNet} based on CNN. For Transformer-based backbones, it is evident that larger networks achieve higher accuracy in address localization. The performance of ResNet-50 is inferior to ViT-B/32 since the former has fewer parameters.
In practice, a balance can be struck according to computational resources and performance requirements.

\subsection{Qualitative Results}
\label{sec:visulization}
Since the address embedding gets inherent alignment with the image feature in our AddressCLIP, we can not only provide precise addresses for query images but also estimate the distribution of images in geographic space according to query addresses. In~\cref{fig:visualization}, we display the embedding similarity distribution in the map of Pittsburgh between images and given address queries. Specifically, Fig.~\ref{fig:visualization} (a) and (e) show the similarity map of two address queries at the neighborhood level, \ie, Downtown and Uptown. The area covered by the highlighted part is consistent with the actual block in the map. Similarly, the results of street-level querying are depicted in~\cref{fig:visualization} (b), (c), (f), and (g). Our semantic partition strategy further enables AddressCLIP to distinguish sub-street level regions within the same street, Forbes Avenue, as shown in~\cref{fig:visualization} (d) and (h).

\begin{table}
  \begin{minipage}{0.99\textwidth}
\scriptsize
\centering  
\topcaption{Comparison examples between representative multimodal large language models and our LLaVA-IAL for the IAL task in Pittsburgh.}
\scalebox{0.88}{
\begin{tabular}{l | p{5.4cm} | p{5.4cm} }
\toprule
 \multicolumn{3}{l}{\bf Visual input examples for Image Address Localization:}  \\
\midrule

&  \includegraphics[height=1.6cm]{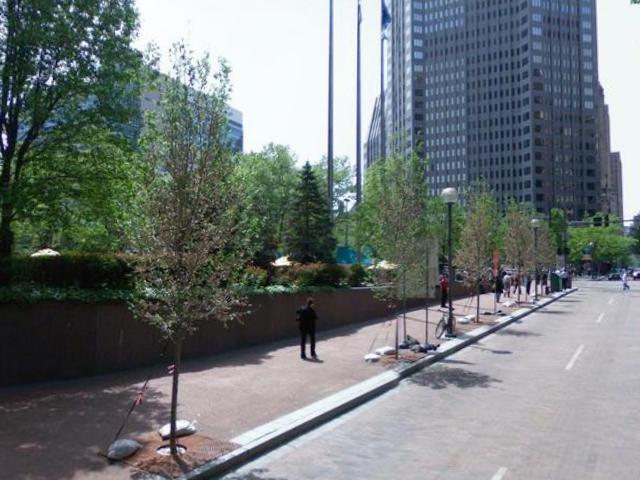} &  \includegraphics[height=1.6cm]{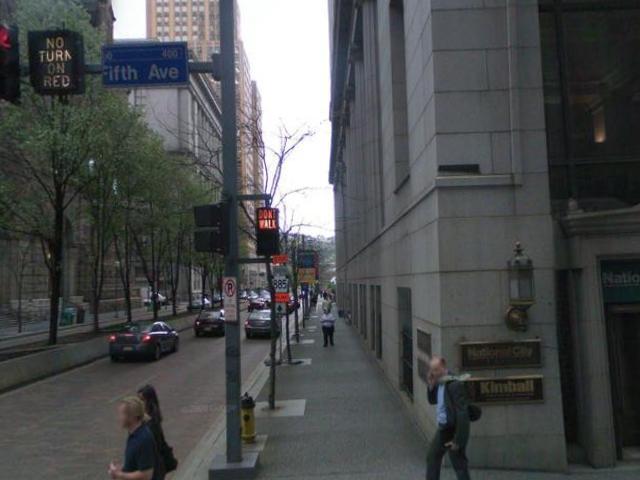} \\
& \textcolor{red}{Grant Street, Downtown} & \textcolor{red}{Fifth Avenue, Downtown} \\
\midrule
User & \multicolumn{2}{l}{Where might this photo have been taken? Tell me its street level address.}\\
\midrule
LLaVA-IAL & The address of this photo might be Grant Street, Downtown, Pittsburgh, PA, USA. & The address of this photo might be Fifth Avenue, Downtown, Pittsburgh, PA, USA.\\
\midrule

GPT-4V~\cite{achiam2023gpt} & This photo was taken in Peavey Plaza in Minneapolis, Minnesota, USA, on the corner of Nicollet Mall and 11th Street. & This photo was taken at 400 Fifth Avenue in Pittsburgh, Pennsylvania. \\

Gemini-Pro-V~\cite{team2023gemini} & This photo was taken on the Rose Fitzgerald Kennedy Greenway in Boston, Massachusetts. & The photo was taken on Fifth Avenue in Pittsburgh, Pennsylvania. \\

QWen-VL~\cite{bai2023qwen} & This photo appears to have been taken on a city street, possibly in a downtown area or commercial district. & This photo appears to have been taken on Fifth Avenue in Pittsburgh, Pennsylvania, USA. The street sign in the image confirms this location.
\\ \bottomrule
\end{tabular}
}
\label{tab:example_bench}  
  \end{minipage}
\end{table}

\subsection{Comparison with "Image-GPS-Address" Pipeline}
\label{sec:exp_diss}
We select state-of-the-art retrieval-based image geo-localization methods~\cite{CosPlace,MixVPR,EigenPlaces,AnyLoc,SALAD} to predict GPS coordinates from a given image, followed by the reverse Geocoding API to obtain a readable textual address. More details about the "Image-GPS-Address" pipeline are provided in the appendix. 
Tab.~\ref{tab:time_storage_comparison} shows the comparison results of AddressCLIP and VPR methods~\cite{TransVPR,r2former,SALAD} in terms of storage space and time overhead. Our AddressCLIP does not require database storage space or retrieval and reranking time consumption, indicating the high efficiency of IAL methods compared to VPR methods. Tab.~\ref{tab:retieval_with_api} shows their performance comparisons on the Pitts-IAL dataset, where all models adopt ResNet-50 as the backbone except for AnyLoc~\cite{AnyLoc} and SALAD~\cite{SALAD} with larger pretrained DINOv2~\cite{dinov2}.
As can be seen, AddressCLIP surpasses the performance of the "Image-GPS-Address" based methods, which indicates that the compounded error of retrieval methods and Geocoding makes the "Image-GPS-Address" pipeline less effective than an end-to-end address localization method.

\subsection{Limitations and Future Work}
The proposed AddressCLIP can be regarded as a discriminative model, limited by the set of candidate addresses at inference. In contrast, generative models such as multimodal large language models (MLLMs)~\cite{achiam2023gpt,team2023gemini,bai2023qwen} can yield more flexible and interactive geographic textual information but may not offer precise administrative addresses. In future work, we plan to explore the potential of MLLMs for the IAL task. To this end, we have made an attempt by constructing a question-and-answer dataset based on Pitts-IAL and adopting the LLaVA-1.5~\cite{liu2023improved} model for instruct tuning. Tab.~\ref{tab:example_bench} shows some examples comparing with representative MLLMs. It is evident that LLaVA-IAL, fine-tuned with instructions, can predict accurate administrative addresses consistent with address hierarchy, while other compared MLLMs are unable to predict addresses without landmarks or street signs and lack standardized output formats.

\section{Conclusion}
In this study, we introduce the problem of image address localization and propose three IAL datasets for evaluation and subsequent research.
To facilitate the alignment of images and addresses for tackling the problem, we propose the AddressCLIP framework consisting of image-text alignment and image-geography matching. Extensive experiments on the proposed datasets validate that our method outperforms transfer learning methods that transfer CLIP to downstream tasks. We compare the proposed method with the existing two-stage address localization pipeline based on the image geo-localization technology and discuss AddressCLIP's application in real-world situations. Finally, we explore the potential of multimodal large language models for address localization.

\section*{Acknowledgements}
This work was supported by the National Natural Science Foundations of China (Grants No.62376267, 62076242) and the innoHK project.

% \clearpage\mbox{}Page \thepage\ of the manuscript.
% \clearpage\mbox{}Page \thepage\ of the manuscript.
% \clearpage\mbox{}Page \thepage\ of the manuscript.
% \clearpage\mbox{}Page \thepage\ of the manuscript.
% \clearpage\mbox{}Page \thepage\ of the manuscript. This is the last page.
% Now we have reached the maximum length of an ECCV \ECCVyear{} submission (excluding references).
% References should start immediately after the main text, but can continue past p.\ 14 if needed.
% \clearpage  % TODO REVIEW/FINAL: This \clearpage needs to be removed from both review and camera-ready versions.

\bibliographystyle{splncs04}
\bibliography{egbib}

\begin{thebibliography}{10}
\providecommand{\url}[1]{\texttt{#1}}
\providecommand{\urlprefix}{URL }
\providecommand{\doi}[1]{https://doi.org/#1}

\bibitem{achiam2023gpt}
Achiam, J., Adler, S., Agarwal, S., Ahmad, L., Akkaya, I., Aleman, F.L., Almeida, D., Altenschmidt, J., Altman, S., Anadkat, S., et~al.: Gpt-4 technical report. arXiv preprint arXiv:2303.08774  (2023)

\bibitem{GSV-cities}
Ali-bey, A., Chaib-draa, B., Gigu{\`e}re, P.: Gsv-cities: Toward appropriate supervised visual place recognition. Neurocomputing  \textbf{513},  194--203 (2022)

\bibitem{MixVPR}
Ali-Bey, A., Chaib-Draa, B., Giguere, P.: Mixvpr: Feature mixing for visual place recognition. In: WACV. pp. 2998--3007 (2023)

\bibitem{NetVLAD}
Arandjelovic, R., Gronat, P., Torii, A., Pajdla, T., Sivic, J.: Netvlad: Cnn architecture for weakly supervised place recognition. In: CVPR. pp. 5297--5307 (2016)

\bibitem{bai2023qwen}
Bai, J., Bai, S., Yang, S., Wang, S., Tan, S., Wang, P., Lin, J., Zhou, C., Zhou, J.: Qwen-vl: A frontier large vision-language model with versatile abilities. arXiv preprint arXiv:2308.12966  (2023)

\bibitem{Surf}
Bay, H., Ess, A., Tuytelaars, T., Van~Gool, L.: Speeded-up robust features (surf). Computer vision and image understanding  \textbf{110}(3),  346--359 (2008)

\bibitem{CosPlace}
Berton, G., Masone, C., Caputo, B.: Rethinking visual geo-localization for large-scale applications. In: CVPR. pp. 4878--4888 (2022)

\bibitem{EigenPlaces}
Berton, G., Trivigno, G., Caputo, B., Masone, C.: Eigenplaces: Training viewpoint robust models for visual place recognition. In: Proceedings of the IEEE/CVF International Conference on Computer Vision (ICCV). pp. 11080--11090 (October 2023)

\bibitem{improve_vlad1}
Berton, G.M., Paolicelli, V., Masone, C., Caputo, B.: Adaptive-attentive geolocalization from few queries: A hybrid approach. In: WACV. pp. 2918--2927 (2021)

\bibitem{GeoCLIP}
Cepeda, V.V., Nayak, G.K., Shah, M.: Geoclip: Clip-inspired alignment between locations and images for effective worldwide geo-localization. arXiv preprint arXiv:2309.16020  (2023)

\bibitem{chen2020simple}
Chen, T., Kornblith, S., Norouzi, M., Hinton, G.: A simple framework for contrastive learning of visual representations. In: ICML. pp. 1597--1607 (2020)

\bibitem{hierarachies}
Clark, B., Kerrigan, A., Kulkarni, P.P., Cepeda, V.V., Shah, M.: Where we are and what we're looking at: Query based worldwide image geo-localization using hierarchies and scenes. In: CVPR. pp. 23182--23190 (2023)

\bibitem{vpr2004}
Csurka, G., Dance, C., Fan, L., Willamowski, J., Bray, C.: Visual categorization with bags of keypoints. In: ECCV Workshop. vol.~1, pp.~1--2 (2004)

\bibitem{ViT}
Dosovitskiy, A., Beyer, L., Kolesnikov, A., Weissenborn, D., Zhai, X., Unterthiner, T., Dehghani, M., Minderer, M., Heigold, G., Gelly, S., et~al.: An image is worth 16x16 words: Transformers for image recognition at scale. arXiv preprint arXiv:2010.11929  (2020)

\bibitem{gao2023clip}
Gao, P., Geng, S., Zhang, R., Ma, T., Fang, R., Zhang, Y., Li, H., Qiao, Y.: Clip-adapter: Better vision-language models with feature adapters. International Journal of Computer Vision pp. 1--15 (2023)

\bibitem{improve_vlad2}
Ge, Y., Wang, H., Zhu, F., Zhao, R., Li, H.: Self-supervising fine-grained region similarities for large-scale image localization. In: ECCV. pp. 369--386 (2020)

\bibitem{R-MAC}
Gordo, A., Almazan, J., Revaud, J., Larlus, D.: End-to-end learning of deep visual representations for image retrieval. International Journal of Computer Vision  \textbf{124}(2),  237--254 (2017)

\bibitem{StreetCLIP}
Haas, L., Alberti, S., Skreta, M.: Learning generalized zero-shot learners for open-domain image geolocalization. arXiv preprint arXiv:2302.00275  (2023)

\bibitem{patch-netvlad}
Hausler, S., Garg, S., Xu, M., Milford, M., Fischer, T.: Patch-netvlad: Multi-scale fusion of locally-global descriptors for place recognition. In: CVPR. pp. 14141--14152 (2021)

\bibitem{he2020momentum}
He, K., Fan, H., Wu, Y., Xie, S., Girshick, R.: Momentum contrast for unsupervised visual representation learning. In: CVPR. pp. 9729--9738 (2020)

\bibitem{ResNet}
He, K., Zhang, X., Ren, S., Sun, J.: Deep residual learning for image recognition. In: CVPR. pp. 770--778 (2016)

\bibitem{sota}
Izquierdo, S., Civera, J.: Optimal transport aggregation for visual place recognition. arXiv preprint arXiv:2311.15937  (2023)

\bibitem{SALAD}
Izquierdo, S., Civera, J.: Optimal transport aggregation for visual place recognition. arXiv preprint arXiv:2311.15937  (2023)

\bibitem{vpr2008}
Jegou, H., Douze, M., Schmid, C.: Hamming embedding and weak geometric consistency for large scale image search. In: ECCV. pp. 304--317 (2008)

\bibitem{VLAD}
J{\'e}gou, H., Perronnin, F., Douze, M., S{\'a}nchez, J., P{\'e}rez, P., Schmid, C.: Aggregating local image descriptors into compact codes. IEEE transactions on pattern analysis and machine intelligence  \textbf{34}(9),  1704--1716 (2011)

\bibitem{jia2021scaling}
Jia, C., Yang, Y., Xia, Y., Chen, Y.T., Parekh, Z., Pham, H., Le, Q., Sung, Y.H., Li, Z., Duerig, T.: Scaling up visual and vision-language representation learning with noisy text supervision. In: ICML. pp. 4904--4916 (2021)

\bibitem{CRN}
Jin~Kim, H., Dunn, E., Frahm, J.M.: Learned contextual feature reweighting for image geo-localization. In: CVPR. pp. 2136--2145 (2017)

\bibitem{AnyLoc}
Keetha, N., Mishra, A., Karhade, J., Jatavallabhula, K.M., Scherer, S., Krishna, M., Garg, S.: Anyloc: Towards universal visual place recognition. IEEE Robotics and Automation Letters  (2023)

\bibitem{khattak2023maple}
Khattak, M.U., Rasheed, H., Maaz, M., Khan, S., Khan, F.S.: Maple: Multi-modal prompt learning. In: CVPR. pp. 19113--19122 (2023)

\bibitem{li2022blip}
Li, J., Li, D., Xiong, C., Hoi, S.: Blip: Bootstrapping language-image pre-training for unified vision-language understanding and generation. In: ICML. pp. 12888--12900 (2022)

\bibitem{liu2023improved}
Liu, H., Li, C., Li, Y., Lee, Y.J.: Improved baselines with visual instruction tuning. arXiv preprint arXiv:2310.03744  (2023)

\bibitem{LLaVA}
Liu, H., Li, C., Wu, Q., Lee, Y.J.: Visual instruction tuning. ArXiv  \textbf{abs/2304.08485} (2023), \url{https://api.semanticscholar.org/CorpusID:258179774}

\bibitem{SFRS}
Liu, L., Li, H., Dai, Y.: Stochastic attraction-repulsion embedding for large scale image localization. In: ICCV. pp. 2570--2579 (2019)

\bibitem{local2004}
Lowe, D.G.: Distinctive image features from scale-invariant keypoints. International journal of computer vision  \textbf{60},  91--110 (2004)

\bibitem{global2006}
Oliva, A., Torralba, A.: Building the gist of a scene: The role of global image features in recognition. Progress in brain research  \textbf{155},  23--36 (2006)

\bibitem{dinov2}
Oquab, M., Darcet, T., Moutakanni, T., Vo, H., Szafraniec, M., Khalidov, V., Fernandez, P., Haziza, D., Massa, F., El-Nouby, A., et~al.: Dinov2: Learning robust visual features without supervision. arXiv preprint arXiv:2304.07193  (2023)

\bibitem{GeoLocator}
Pramanick, S., Nowara, E.M., Gleason, J., Castillo, C.D., Chellappa, R.: Where in the world is this image? transformer-based geo-localization in the wild. In: ECCV. pp. 196--215 (2022)

\bibitem{GEM}
Radenovi{\'c}, F., Tolias, G., Chum, O.: Fine-tuning cnn image retrieval with no human annotation. IEEE transactions on pattern analysis and machine intelligence  \textbf{41}(7),  1655--1668 (2018)

\bibitem{CLIP}
Radford, A., Kim, J.W., Hallacy, C., Ramesh, A., Goh, G., Agarwal, S., Sastry, G., Askell, A., Mishkin, P., Clark, J., et~al.: Learning transferable visual models from natural language supervision. In: ICML. pp. 8748--8763 (2021)

\bibitem{radford2021learning}
Radford, A., Kim, J.W., Hallacy, C., Ramesh, A., Goh, G., Agarwal, S., Sastry, G., Askell, A., Mishkin, P., Clark, J., et~al.: Learning transferable visual models from natural language supervision. In: ICML. pp. 8748--8763 (2021)

\bibitem{vpr2007}
Schindler, G., Brown, M., Szeliski, R.: City-scale location recognition. In: CVPR. pp.~1--7. IEEE (2007)

\bibitem{LAION-5B}
Schuhmann, C., Beaumont, R., Vencu, R., Gordon, C., Wightman, R., Cherti, M., Coombes, T., Katta, A., Mullis, C., Wortsman, M., Schramowski, P., Kundurthy, S., Crowson, K., Schmidt, L., Kaczmarczyk, R., Jitsev, J.: Laion-5b: An open large-scale dataset for training next generation image-text models. ArXiv  \textbf{abs/2210.08402} (2022), \url{https://api.semanticscholar.org/CorpusID:252917726}

\bibitem{cplanet}
Seo, P.H., Weyand, T., Sim, J., Han, B.: Cplanet: Enhancing image geolocalization by combinatorial partitioning of maps. In: ECCV. pp. 536--551 (2018)

\bibitem{VGG}
Simonyan, K., Zisserman, A.: Very deep convolutional networks for large-scale image recognition. arXiv preprint arXiv:1409.1556  (2014)

\bibitem{team2023gemini}
Team, G., Anil, R., Borgeaud, S., Wu, Y., Alayrac, J.B., Yu, J., Soricut, R., Schalkwyk, J., Dai, A.M., Hauth, A., et~al.: Gemini: a family of highly capable multimodal models. arXiv preprint arXiv:2312.11805  (2023)

\bibitem{pitts}
Torii, A., Sivic, J., Pajdla, T., Okutomi, M.: Visual place recognition with repetitive structures. In: CVPR. pp. 883--890 (2013)

\bibitem{divide&classify}
Trivigno, G., Berton, G., Aragon, J., Caputo, B., Masone, C.: Divide\&classify: Fine-grained classification for city-wide visual geo-localization. In: ICCV. pp. 11142--11152 (2023)

\bibitem{vaswani2017attention}
Vaswani, A., Shazeer, N., Parmar, N., Uszkoreit, J., Jones, L., Gomez, A.N., Kaiser, {\L}., Polosukhin, I.: Attention is all you need. NeurIPS  \textbf{30} (2017)

\bibitem{TransVPR}
Wang, R., Shen, Y., Zuo, W., Zhou, S., Zheng, N.: Transvpr: Transformer-based place recognition with multi-level attention aggregation. In: CVPR. pp. 13648--13657 (2022)

\bibitem{planet}
Weyand, T., Kostrikov, I., Philbin, J.: Planet-photo geolocation with convolutional neural networks. In: ECCV. pp. 37--55 (2016)

\bibitem{wilson2023image}
Wilson, D., Zhang, X., Sultani, W., Wshah, S.: Image and object geo-localization. International Journal of Computer Vision pp. 1--43 (2023)

\bibitem{zhou2022conditional}
Zhou, K., Yang, J., Loy, C.C., Liu, Z.: Conditional prompt learning for vision-language models. In: CVPR. pp. 16816--16825 (2022)

\bibitem{zhou2022learning}
Zhou, K., Yang, J., Loy, C.C., Liu, Z.: Learning to prompt for vision-language models. International Journal of Computer Vision  \textbf{130}(9),  2337--2348 (2022)

\bibitem{r2former}
Zhu, S., Yang, L., Chen, C., Shah, M., Shen, X., Wang, H.: R2former: Unified retrieval and reranking transformer for place recognition. In: CVPR. pp. 19370--19380 (2023)

\bibitem{APANet}
Zhu, Y., Wang, J., Xie, L., Zheng, L.: Attention-based pyramid aggregation network for visual place recognition. In: ACM MM. pp. 99--107 (2018)

\end{thebibliography}

\appendix
\section*{Appendix}
\section{Dataset Details}
We provide more details of the dataset construction and statistical results.

\subsection{Return Information of Google Maps API}
For a querying pair of GPS coordinates (\emph{latitude, longitude}), the reverse Geocoding API of Google Maps will return a list of address results, which are sorted by the distance between the address and query GPS coordinates. The information contained in each returned result is shown in~\cref{fig:result_info}. We also provide some examples of formatted addresses in returned results of the same query in~\cref{fig:result_address}. 

Apart from the formatted address information, for each result, location type information is also provided to mark what kind of address is returned. Specifically, "ROOFTOP" means the address is an accurate location (usually a building). "RANGE\_INTERPOLATED" means the result is an approximate position (usually on the road). "GEOMETRIC\_CENTER" means the result is the geometric center of a multi-segment line (such as a street) or a polygon (such as an area). "APPROXIMATE" indicates that the returned result is an approximate location. We only use the formatted address and location type for address annotation. More details can be found in the \href{https://developers.google.cn/maps/documentation/javascript/geocoding}{official documentation} of the Geocoding API of Google Maps.

\subsection{Details of Address Annotation}
We annotate administrative address information from coarse to fine for images with GPS coordinates in image geo-localization datasets by reverse geocoding, address extraction, and semantic address partition. An detailed illustration is shown in Fig.~\ref{fig:annotation}.

\noindent\textbf{\emph{Reverse Geocoding.}} Reverse geocoding is also known as address lookup, which converts a location into an administrative address that is easy to understand. We use the reverse Geocoding API of Google Maps to obtain the address information for each location. Specifically, given the GPS coordinates $(latitude, longitude)$ of a location, the reverse Geocoding API returns a list of addresses that are ordered by their match degree with the coordinate, \eg $[A^{(1)}, A^{(2)}, \cdots, A^{(R)}]$, together with their location types. However, simply selecting $A^{(1)}$ as the ground truth address is often imprecise since the API tends to match the GPS coordinates to a place’s or building’s geometry center and return its address. For instance, when a large building is situated at an intersection of \textit{Street A} and \textit{Street B} but faces \textit{Street A}, the coordinates on \textit{Street B} near this building might always be labeled as \textit{Street A} with $A^{(1)}$.

\begin{figure}[h]
    \centering
\begin{minipage}[t]{0.48\textwidth}
\centering
\begin{minted}[fontsize=\tiny]{python}
results[]: {types[]: string, formatted_address: string,
address_components[]: {short_name: string,
long_name: string,
postcode_localities[]: string, types[]: string},
partial_match: boolean, place_id: string,
postcode_localities[]: string,
geometry: {location: LatLng,
location_type: GeocoderLocationType
viewport: LatLngBounds, bounds: LatLngBounds}}
\end{minted}
\caption{The specific information and their types in each returned result.}
\label{fig:result_info}
\end{minipage}
\begin{minipage}[t]{0.48\textwidth}
\centering
\begin{minted}[fontsize=\tiny]{python}
results[0].formatted_address: "277 Bedford Ave, Brooklyn, NY, USA"
results[1].formatted_address: "Grand St/Bedford Av, Brooklyn, NY, USA"
results[2].formatted_address: "Williamsburg, Brooklyn, NY, USA"
results[3].formatted_address: "Brooklyn, NY, USA"
results[4].formatted_address: "New York, NY, USA"
results[5].formatted_address: "Brooklyn, NY, USA"
results[6].formatted_address: "Kings County, NY, USA"
results[7].formatted_address: "New York Metropolitan Area, USA"
results[8].formatted_address: "New York, USA"
        \end{minted}
\caption{Examples of formatted addresses in the returned results of the same query.}
\label{fig:result_address}
\end{minipage}
\end{figure}

\begin{figure}[h]
    \centering
    \includegraphics[width=1.0\linewidth]{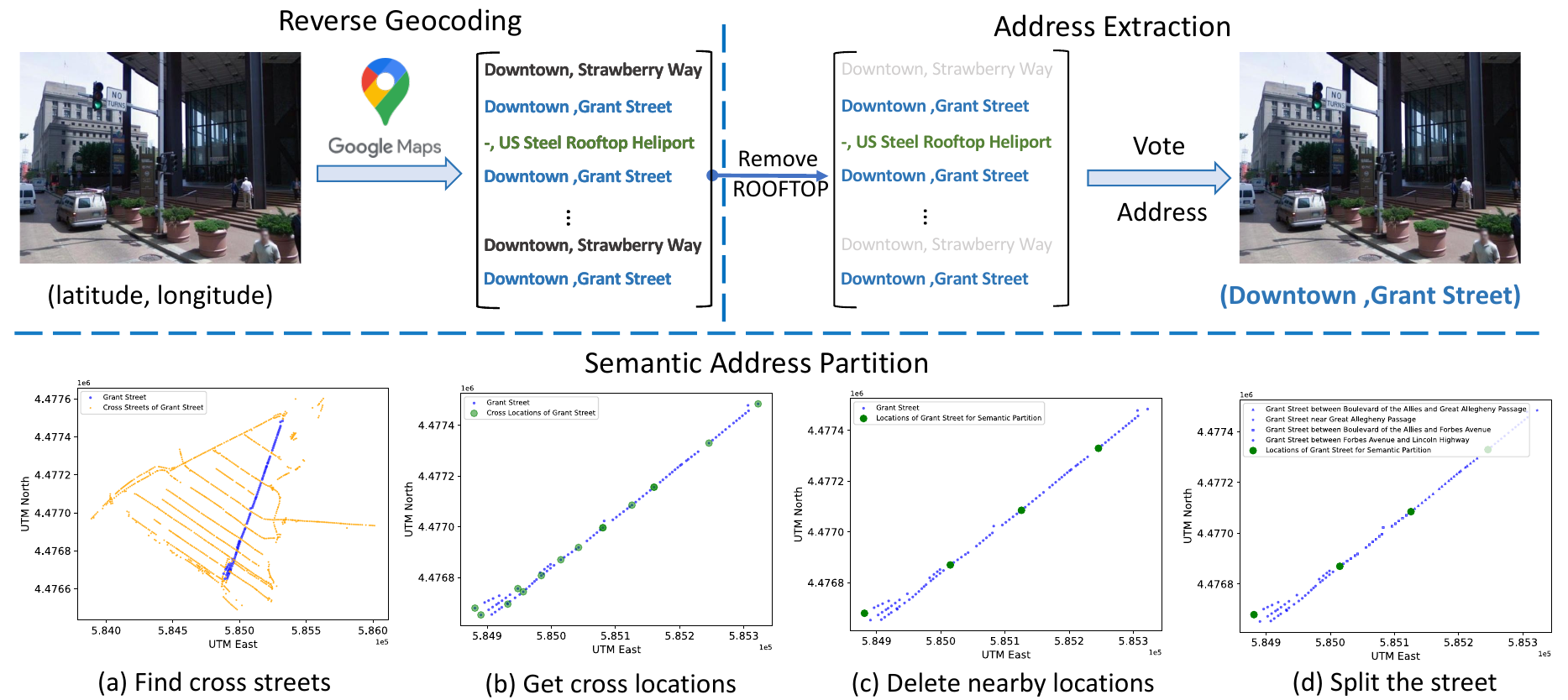}
    \caption{The pipeline of address annotation, including the reverse Geocoding from GPS to addresses, extraction of address information, and semantic address partition.}
    \label{fig:annotation}
\end{figure}

Note that accessing the Google Maps API incurs certain costs, and for large and dense datasets like SF-IAL, which covers hundreds of thousands of locations, retrieving the address for every single location is prohibitively expensive and unaffordable for us. Therefore, for the SF-IAL dataset, we randomly sampled a subset of locations to conduct reverse geocoding, while the information for the remaining locations was filled in using the nearest neighbor's information. Therefore, we require additional post-processing steps to ensure the accuracy of the street-level address annotation.

\noindent\textbf{\emph{Address Extraction.}} To alleviate the above issue, we adopt the following three steps to extract accurate street-level address information from the returned list for each location. 
Firstly, we remove the returned addresses that are matched to buildings along the street, whose location type is marked as "ROOFTOP" in the results. Secondly, among the remaining addresses, we choose the most frequently occurring address as the definitive address through a voting mechanism. Thirdly, we correct inaccurately labeled addresses through manual random verification.
Building upon the aforementioned efforts, we have thus obtained accurate street-level annotations for each location. This serves as a vital foundation for the subsequent sub-street segmentation.

\begin{table}[tb]
  \caption{More statistics of the proposed Image Address Localization datasets.}
  \label{tab:dataset}
  \centering
  \scriptsize
  \tabcolsep=0.1cm
  \begin{tabular}{l | c | c | c | c | c}
    \toprule
    Dataset & \makecell[c]{Covered \\ Area} & \makecell[c]{\# Neighborhood \\ train / test} & \makecell[c]{\# Street \\ train / test} & \makecell[c]{\# Sub-Street \\ train / test} & \makecell[c]{\# Locations \\ Returned by API / Total}\\
    \midrule
    Pitts-IAL & 20 km$^2$ & 19 / 19 & 194 / 165 & 428 / 327 & 10,586 / 10,586\\
    SF-IAL-Base & 6 km$^2$ & 15 / 15 & 121 / 110 & 400 / 369 & 8,371 / 17,067\\
    SF-IAL-Large & 170 km$^2$ &124 / 124 & 332 / 327 & 3,616 / 3,406 & 17,686 / 233,820\\
  \bottomrule
  \end{tabular}
\end{table}

\begin{figure}[tb]
    \centering
    \includegraphics[width=1.0\linewidth]{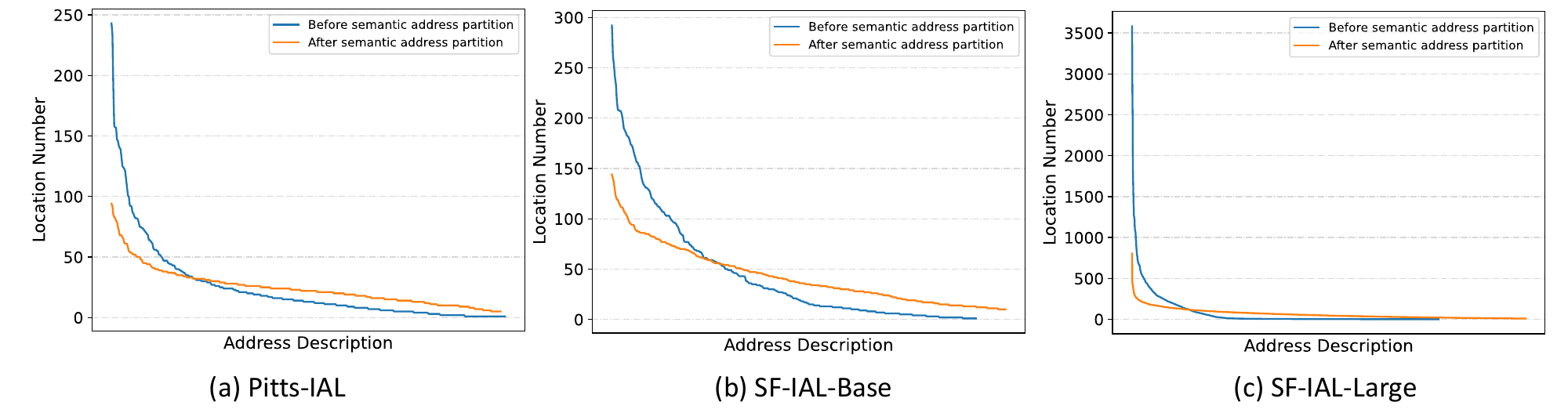}
    \caption{The address distribution in the three Image Address Localization datasets. Both the distributions before and after semantic address partition are demonstrated.}
    \label{fig:distribution}
\end{figure}

\noindent\textbf{\emph{Semantic Address Partition.}}
As mentioned in the main paper, to balance the length of streets while eliminating the naming ambiguity at intersections, we adopt the semantic address partition strategy for a more granular segmentation of streets. Specifically, the process is as follows: First, identify all the intersecting streets (\textcolor{orange}{orange}) for each main street (\textcolor{blue}{blue}). Second, obtain the intersection points (\textcolor{green}{green} points) on the main street that intersect with other streets. Third, remove closely spaced intersection points based on a certain threshold distance to avoid redundancy and too short sub-streets. Finally, split the main street into sub-streets and assign their names based on the remaining intersection points. In this way, the textual representation of addresses consists of the main street name and the name of one or two streets that intersect it. Moreover, to mitigate the long-tail issue inherent in street distributions, we employ a method of clustering adjacent addresses, merging shorter sub-streets ($< 5$ locations) into longer ones or broader areas. This process ensures that the description of addresses is both general and precise. It is important for the datasets that have irregular streets and sparse locations, \eg, Pitts-IAL.

\begin{figure}[tb]
    \centering
    \includegraphics[width=1.0\linewidth]{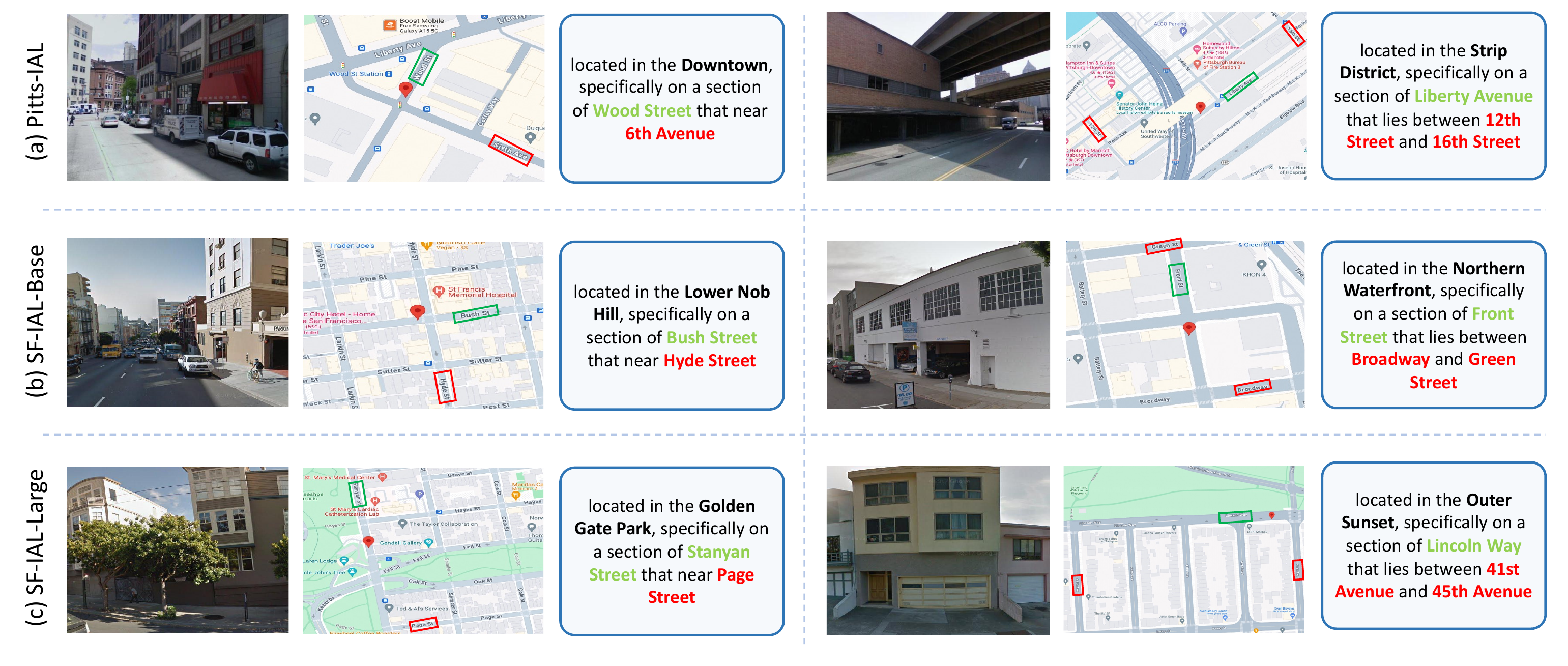}
    \caption{Examples of address annotation for the three Image Address Localization datasets and their locations on the map.}
    \label{fig:dataset_exp}
\end{figure}

\subsection{More Statistics and Visualizations}
We have presented the basic information of the three image address localization datasets in Sec. \textcolor{red}{5.2} of the main paper. In Tab.~\ref{tab:dataset}, we provide more statistical details of the introduced three datasets, including the geographic area coverage of the locations, the number of neighborhoods, streets, and sub-streets in the training and test sets, as well as the number of addresses captured by the Google Maps API. 

Additionally, in Fig.~\ref{fig:distribution}, we illustrate the distribution of the number of locations associated with each address across the three datasets. The phenomenon of long-tail distribution is significantly mitigated after applying the semantic address partition strategy, resulting in a more balanced distribution of the number of locations per address. 

Fig.~\ref{fig:dataset_exp} showcases examples of address annotations along with their corresponding positions on the map from three introduced datasets.

\section{More Implementation Details}
In training, the Adam ($\beta_1=0.9$ $\beta_2=0.98$) is adopted as the optimizer with the cosine learning rate from 2.4e-5 to 2.4e-8. We set the loss weights $\alpha$, $\beta$ and $\gamma$ of the final objective to 1, 0.2, and 0.8, respectively. The batch size is set to 32 for each GPU and all the model is trained for 100 epochs on 8 Tesla V100.

\section{Additional Ablations about Scene Caption}
\subsection{Scene Caption with Different Models}
In this section, we present and discuss the impact of scene captions generated by different models. We utilize different versions of the vision-language model BLIP~\cite{li2022blip} for image captioning, namely BLIP-Caption-Base and BLIP-Caption-Large, to generate additional scene descriptions. Both of the two models are prompted with "A street view of". The minimum and maximum length of the output captions are set to 10 and 30, respectively.

\begin{table}
\caption{Comparison examples between the generated captions by BLIP-Caption-Base and BLIP-Caption-Large.}
  \begin{minipage}{0.99\textwidth}
\scriptsize
\centering  
\scalebox{0.88}{
\begin{tabular}{l | p{3.1cm} | p{3.1cm} | p{3.1cm} | p{3.1cm}}
\toprule
 \multicolumn{5}{l}{\bf Image Captioning Examples with BLIP~\cite{li2022blip}:}  \\
\midrule
&  \includegraphics[height=2.1cm]{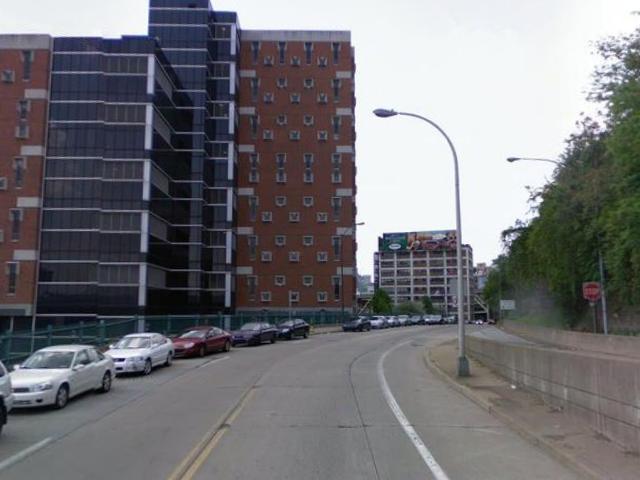} &  \includegraphics[height=2.1cm]{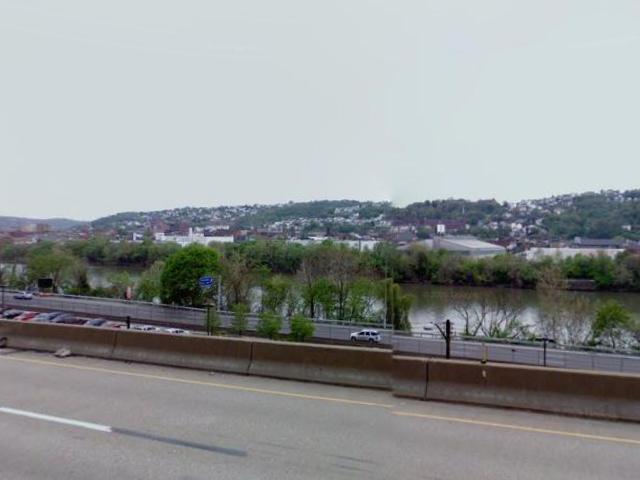} & \includegraphics[height=2.1cm]{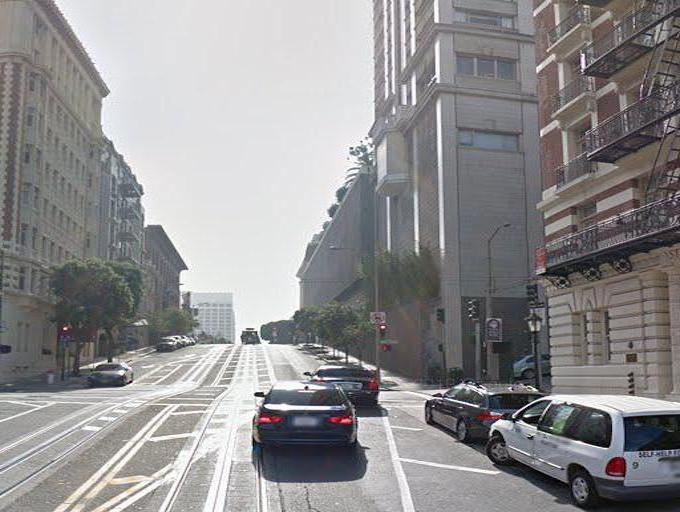} & \includegraphics[height=2.1cm]{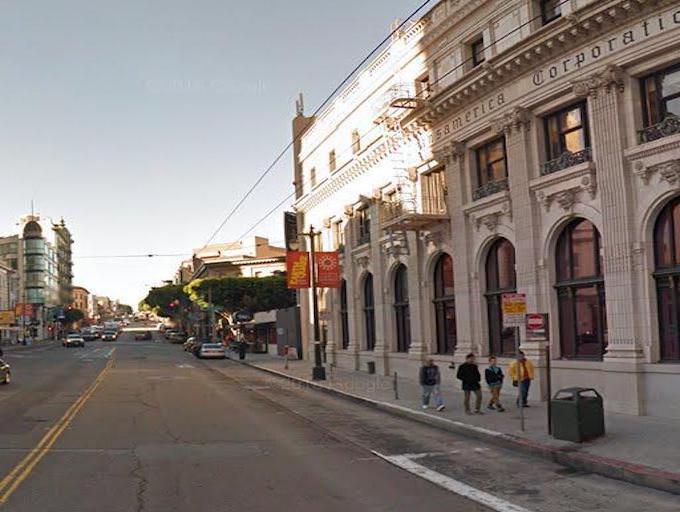}\\
\midrule
Base & a street view of a large building with cars parked on the side & a street view of a river and a city & a street view of a city with cars and buildings & a street view of a city with buildings and people \\
\midrule
Large & a street view of a city with a lot of cars parked on the side of the road and tall buildings & a street view of a river and a city with a bridge in the background and a car driving on the road & a street view of a city street with cars and buildings on both sides of the street and a tram crossing & a street view of a city street with a few people walking on the sidewalk and a building in the background \\
\bottomrule
\end{tabular}
}
\label{tab:example_blip}  
  \end{minipage}
\end{table}

\begin{table}[!htbp]
\begin{center}
\caption{Examples of scene captions of BLIP-Caption-Base and BLIP-Caption-Large models and their performance comparisons on the Pitts-IAL and SF-IAL-Base datasets.}
\scriptsize
\label{tab:scene_caption_model}
\setlength{\tabcolsep}{1.8mm}{
\begin{tabular}{ c  | c  c c c |c c c c }
\toprule
\multirow{2}{*}{Caption by Models} & \multicolumn{4}{c|}{Pitts-IAL} & \multicolumn{4}{c}{SF-IAL-Base} \\
 &  SSA-1 &  SSA-5 & SA-1 &  SA-5 & SSA-1 & SSA-5 & SA-1 & SA-5 \\
 \midrule
BLIP-Caption-Base & 80.28 & 95.99 & 82.48 & 96.43 & 86.25 & 99.00 & 87.41 & 99.19\\
BLIP-Caption-Large & 80.39 & 96.27 & 82.62 & 96.74 & 86.32 & 99.09 & 87.44 & 99.23\\
\bottomrule
\end{tabular}}
\end{center}
\end{table}

Tab.~\ref{tab:example_blip} presents examples of scene captions generated by the two models, where BLIP-Caption-Base produces naive descriptions while BLIP-Caption-Large can generate more elaborate captions. Tab.~\ref{tab:scene_caption_model} shows the results of AddressCLIP on the Pitts-IAL and SF-IAL-Base datasets using different captions generated by the above two models. It can be observed that the resulted two AddressCLIP models achieve comparable performance, with that trained with richer scene captions yielding a slightly higher performance. This suggests that using a more powerful model to generate richer descriptive information can further enhance performance, but the gains might be limited. In practice, one needs to balance the cost of generating scene captions with the performance benefits.

\subsection{Scene Caption Formats}

Tab.~\ref{tab:scene_caption} presents the performance comparison results of different scene caption forms of whether the textual address is incorporated or not. It is clear that the addition of geographical address information improves the accuracy compared to using only scene caption, suggesting that the combination of textual address and scene captions is beneficial in enhancing AddressCLIP's capacity to accurately align images with their corresponding locations.

\begin{table}[!htbp]
\begin{center}
\caption{Performance of different scene caption formats on the proposed datasets.}
\scriptsize
\label{tab:scene_caption}
\setlength{\tabcolsep}{1.2mm}{
\begin{tabular}{ c  | c  c c c |c c c c }
\toprule
\multirow{2}{*}{Scene Caption Format} & \multicolumn{4}{c|}{Pitts-IAL} & \multicolumn{4}{c}{SF-IAL-Base}  \\
 & SSA-1 &  SSA-5 & SA-1 &  SA-5 & SSA-1 & SSA-5 & SA-1 & SA-5 \\
\midrule
scene caption w/o address & 79.32 & 95.51 & 81.44 & 95.87 & 84.79 & 97.99 & 86.04 & 98.19\\
scene caption w/ address & \textbf{80.39} & \textbf{96.27} & \textbf{82.62} & \textbf{96.74} & \textbf{86.32} & \textbf{99.09} & \textbf{87.44} & \textbf{99.23}\\
\bottomrule
\end{tabular}}
\end{center}
\end{table}

\begin{table}[!htbp]
\begin{center}
\scriptsize
\caption{Performance of AddressCLIP on the Pitts-IAL dataset when the granularity of search space is varied. $\mathcal{W}$ is the number of prior streets.}
\label{tab:prior}
\setlength{\tabcolsep}{3.5mm}{
\begin{tabular}{ c | c | c | c c c c  }
\toprule
Settings & None & Neighborhood & $\mathcal{W}$=20 & $\mathcal{W}$=10 & $\mathcal{W}$=5 & $\mathcal{W}$=2 \\
\midrule
SSA-1 & 80.39 & 82.18 & 80.83 & 82.20 & 85.17 & 89.57\\
\bottomrule
\end{tabular}}
\end{center}
\end{table}

\begin{table}[!htbp]
\begin{center}
\caption{Performance of different geographic coverage on the Pitts-IAL dataset.}
\label{tab:geo_overlap}
\scriptsize
\setlength{\tabcolsep}{3.0mm}{
\begin{tabular}{  c |  c | c | c | c }
\toprule
\# Images  & 3 & 6 & 12 & 24 \\
\midrule
SSA-1/SSA-5 & 56.86/84.58 & 69.08/92.06 & 76.80/95.45 & 80.39/96.27 \\
SA-1/SA-5 & 61.12/86.50 & 72.48/93.09 & 79.45/96.01 & 82.62/96.74\\
\bottomrule
\end{tabular}}
\end{center}
\end{table}

\section{Discussion about the Characteristics of AddressCLIP}
\subsection{AddressCLIP with Prior Knowledge}
In practical applications of address localization, users often possess some level of prior geographical context. For instance, while the address of an image may be unknown, the neighborhood or several candidate streets may be known. This additional information can effectively narrow the search space, thereby improving the model’s accuracy due to the reduced number of potential addresses for consideration. To assess AddressCLIP’s adaptability in these situations, we perform experiments where the candidate address is limited within a predefined neighborhood or several streets. Results across different settings are shown in Tab.~\ref{tab:prior}. The performance of AddressCLIP improves as we search from coarse to finer granularity. The model’s capacity to adapt to restricted search spaces affirms its applicability in real-world scenarios where partial geographic context is commonly available.

\subsection{Different Geographic Coverage}
The IAL task assumes that the addresses during testing are covered during training, thus the training sets and testing sets are geographically overlapped. In the main experiments on the Pitts-IAL dataset, 24 images were taken from different perspectives at each location. To explore the potential of the proposed AddressCLIP under conditions with less geographical coverage, we randomly reduce the number of images per location to 12, 6, and 3. Tab.~\ref{tab:geo_overlap} shows the results of different geographic coverage on Pitts-IAL. One can observe that AddressCLIP can preserve \textbf{75\%} of original performance with only \textbf{12.5\%} location coverage, demonstrating its efficiency.

\begin{table}[!htbp]
\begin{center}
\caption{Effect of mixed training on both Pitts-IAL and SF-IAL-Base datasets.}
\label{tab:mix_training}
\scriptsize
\setlength{\tabcolsep}{3.0mm}{
\begin{tabular}{  c |  c  c  | c  c }
\toprule
Train / Test & SSA-1 & SSA-5 & SA-1 & SA-5 \\
\midrule
Pitts / Pitts & 80.39 & 96.27 & 82.62 & 96.74 \\
Pitts + SF / Pitts & 80.46\textcolor{red}{$\bm{_{+0.07}}$} & 95.95\textcolor{red}{$\bm{_{-0.32}}$} & 82.62\textcolor{red}{$\bm{_{-0.00}}$} & 96.51\textcolor{red}{$\bm{_{-0.23}}$} \\
\midrule
SF / SF & 86.32 & 99.09 & 87.44 & 99.23 \\
Pitts + SF / SF & 85.51\textcolor{red}{$\bm{_{-0.81}}$} & 98.31\textcolor{red}{$\bm{_{-0.78}}$} & 86.82\textcolor{red}{$\bm{_{-0.62}}$} & 98.72\textcolor{red}{$\bm{_{-0.51}}$} \\
\midrule
Pitts + SF / Pitts+SF & 83.07 & 97.17 & 84.79 & 97.65 \\
\bottomrule
\end{tabular}}
\end{center}
\end{table}

\subsection{Mixed Training on Multi-city Datasets}
In the main experiments, the proposed AddressCLIP is trained and evaluated on the Pitts-IAL and SF-IAL-Base datasets respectively, but this does not mean that our method can only work on a single city. To explore the potential of AddressCLIP on multiple city datasets, we combine the Pitts-IAL and SF-IAL-Base datasets as a mixed dataset for training. Tab.~\ref{tab:mix_training} shows the performance comparisons with single-city dataset training.
As can be seen, without increasing the model size, the performance of mixed training achieves comparable performance to that trained on each dataset ($\bm{<0.8\%}$ degradation), which shows the scalability and potential of AddressCLIP to work across multiple cities.

\section{Implements of "Image-GPS-Address" Pipeline}
The "Image-GPS-Address" pipeline involves utilizing the image geo-localization technology to predict the GPS coordinates of a given image query and then translating them into textual addresses by the reverse Geocoding API of Google Maps. To compare our proposed end-to-end method with existing solutions, we have implemented this two-stage pipeline for these methods and provided evaluation results on the Pitts-IAL dataset in the Sec. \textcolor{red}{6.5} of the main paper. The experiment demonstrates that our approach achieves superior street-level address localization capabilities, presenting a promising method. We acknowledge that the proposed method is not yet as precise as GPS localization (within 25 meters), but its advantage lies in being an end-to-end solution. Moreover, the predicted textual addresses are more semantically meaningful and align with human description habits.

Specifically, we adopt state-of-the-art geo-localization approaches (\ie, CosPlace~\cite{CosPlace}, MixVPR~\cite{MixVPR}, EigenPlaces~\cite{EigenPlaces}, AnyLoc~\cite{AnyLoc}, SALAD~\cite{SALAD}) and use their publicly available model weights for feature extraction, all of which are claimed to have robust generalization capabilities. The database and query sets of the introduced Pitts-IAL dataset are utilized for the retrieval-based methods. We use ResNet50 with a feature dimension of 512 for image retrieval, except for AnyLoc~\cite{AnyLoc} and SALAD~\cite{SALAD}, which use DINOv2 for feature extraction. For each query image, we first calculate its Euclidean distance with all the database images. Then we select the location of the image that has a minimum distance with the query as the predicted location. This location is used for reverse Geocoding to obtain the textual addresses.

Formally, given a query image $Q$, we define $D$ as the database containing all reference images, where each reference image is denoted as $D_i$, $\forall i \in [{1, 2, ..., N}]$, and $N$ is the total number of images in the database. We compute the Euclidean distance in the feature space between the query image and each reference image in the database as follows:

\begin{equation}
    Dist(Q, D_i) = \sqrt{\sum_{j=1}^{M} (Q(j) - D_i(j))^2},
\end{equation}
where $M$ represents the dimension of the feature. $Q(j)$  and $D_i(j)$ are the $j$-th feature of the query and database image, respectively. Then the predicted location $L_Q$ for the query image is determined by assigning the GPS of the reference image with the minimum Euclidean distance in feature space, \ie,
\begin{equation}
    L_Q = GPS(\operatorname*{argmin}_{D_i \in D} \text{Dist}(Q, D_i)),
\end{equation}
where $GPS(\cdot)$ indicates the lookup table of the Image-GPS pairs in the database.

 In addition, when using the reverse Geocoding API, for fair comparison, we exclude returned address information where the location type is “ROOFTOP”, and choose the most frequently occurring address from the remaining addresses as the final predicted address.

\begin{figure}[tb]
    \centering
    \begin{minted}[fontsize=\scriptsize]{json}
[
  {
    "id": "<path>",
    "image": "<path>",
    "conversations": [
      {
        "from": "human",
        "value": "<image>\nWhere might this photo have been taken? \ 
                    Tell me its street level adress."
      },
      {
        "from": "gpt",
        "value": " The address of this photo might be Grant Street, Pittsburgh, PA."
      },
    ]
  },
  ...
]
\end{minted}
    \caption{An example of the constructed conversation data from the Pitts-IAL dataset.}
    \label{fig:enter-label}
\end{figure}

\section{Details of Instruct Tuning with LLaVA}
Multimodal large language models (MLLMs) are key building blocks for general-purpose visual assistants, and they have become increasingly popular in the research community. To apply MLLMs to the task of image address localization, we construct a multimodal dataset that pairs visual images with textual addresses in a question-and-answer format from Pitts-IAL. Fig.~\ref{fig:enter-label} shows an example. Here, we adopt LLaVA-1.5~\cite{LLaVA} as the MLLM since it demonstrates impressive results on instruction-following and visual reasoning capabilities with open-source code and models. The instruct tuning process involves adjusting the LLaVA model's parameters with LoRA. During training, The AdamW is used as the optimizer and the cosine annealing scheduler is used to adjust the learning rate. We set the batch size to 16 and the learning rate to 1e-4. All training is conducted on 8 GeForce RTX 3090 GPUs with 24GB memory. The training of three epochs costs 20 hours. The input image size is set to 224 × 224. The fine-tuned LLaVA model, LLaVA-IAL, shows a significant improvement in the ability to predict textual addresses from images, which is indicative of its enhanced understanding of the visual and textual cues pertinent to the task of image address localization. This advancement holds promise for applications that require intelligent navigation and seamless interaction between the digital and physical realms.

\begin{figure}[h]
    \centering
    \includegraphics[width=1.0\linewidth]{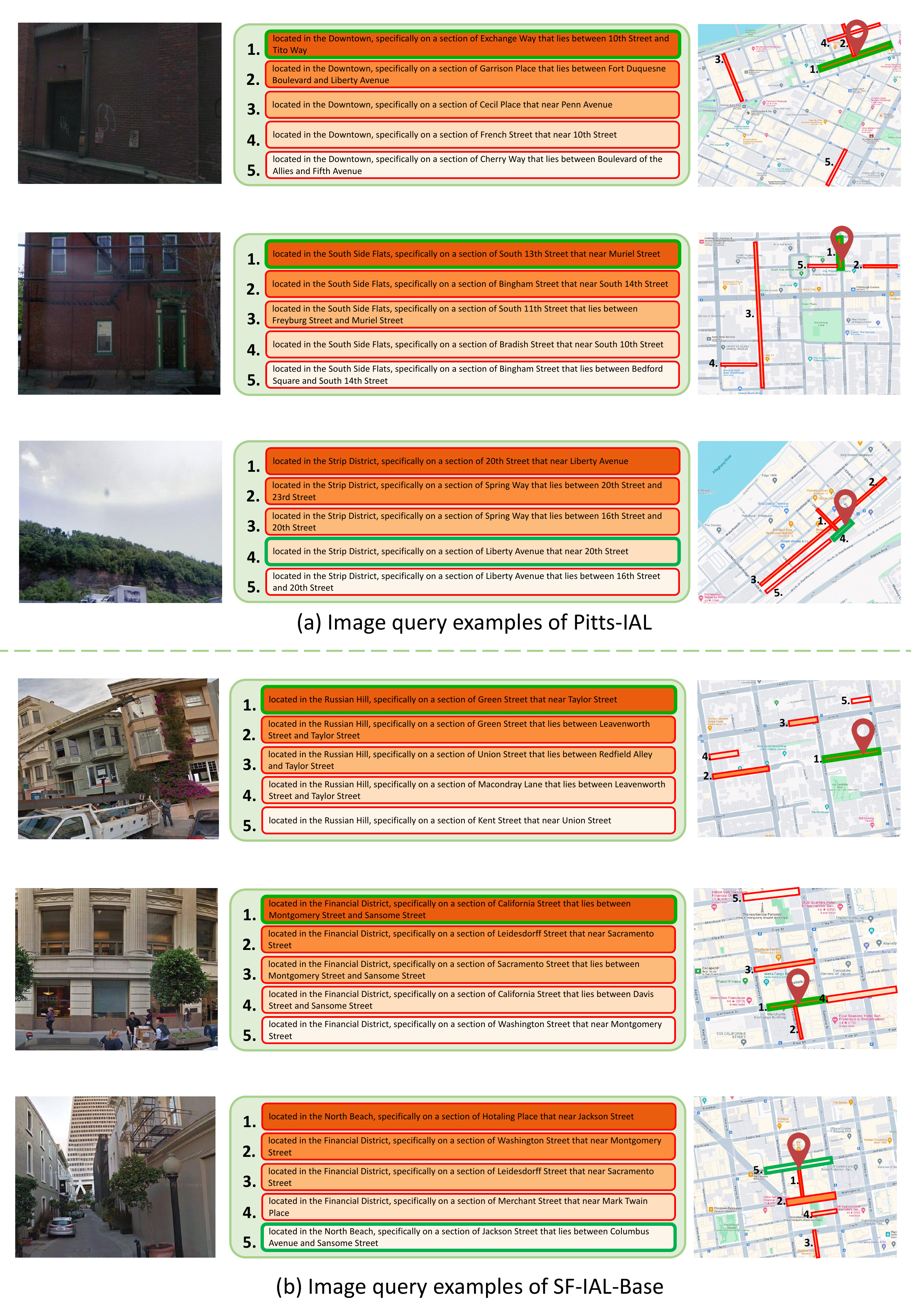}
    \caption{The address localization results predicted by AddressCLIP and their positions on the map according to image queries. The results from Top-1 to Top-5 are displayed, with green boxes indicating correctly predicted addresses and red boxes indicating incorrectly predicted addresses.}
    \label{fig:image_queries}
\end{figure}

\begin{figure}[h]
    \centering
    \includegraphics[width=1.0\linewidth]{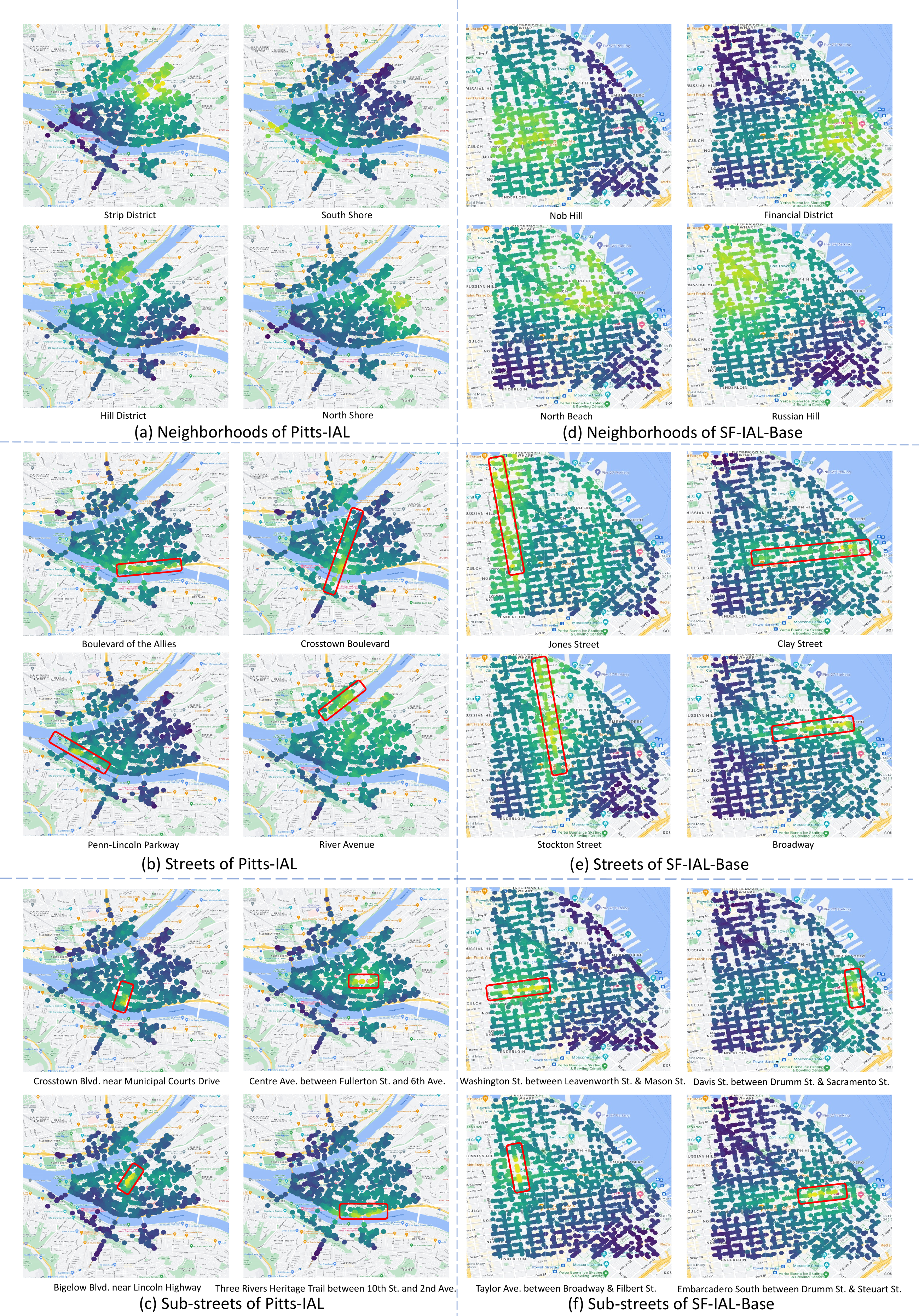}
    \caption{More qualitative demonstrations with a given textual address query using AddressCLIP in Pittsburgh and San Francisco. The brighter the scatter point, the higher the similarity of the embedding between the image and the query address text. The red box represents the actual geographic range of the query street in the map.}
    \label{fig:text_queries}
\end{figure}

\section{Qualitative Demonstration}
In this section, we qualitatively demonstrate the effectiveness of our method. We first show the results of AddressCLIP with the image query. Then, we provide more visualizations of the similarity map between the image embedding and the address text query in Pittsburgh and San Francisco. 

\subsection{AddressCLIP with Image Query}
Fig.~\ref{fig:image_queries} shows the Top-5 textual address predictions generated by the proposed AddressCLIP, based on given image queries, along with their locations on the map. The examples provided come from the Pitts-IAL and SF-IAL-Base datasets. In the majority of cases, the correct prediction is identified within the first address (Top-1), demonstrating AddressCLIP's precise address localization capability. Subsequent predicted addresses are also close to the correct location. Additionally, we showcase some failure examples where the Top-1 prediction is not correct. Even so, the correct address can still be predicted within the Top-5 addresses, and the Top-1 predicted address is typically close to the actual location.

\subsection{AddressCLIP with Address Text Query}
In Fig.~\ref{fig:text_queries}, we display more visualizations of the embedding similarity distribution between images and given address queries on the map of Pittsburgh and San Francisco. It is observed that on both Pitts-IAL and SF-IAL-Base datasets when provided with a text query, our AddressCLIP is capable of effectively pinpointing the approximate area corresponding to the text based on the features of the street view images. The results presented are divided into three levels: neighborhood, street, and sub-street from the top row to the bottom.

\section{Broader Impacts}
In this study, we introduce the problem of image address localization, which aims to predict the textual address where a given image was taken, consistent with how humans typically describe addresses. With the proposed AddressCLIP, we can obtain more semantic address information, which has the potential to revolutionize the way we navigate and interact with physical spaces. The introduced image address localization datasets are derived from open-source datasets Pitts-250k~\cite{NetVLAD} and SF-XL~\cite{CosPlace} as well as publicly available Google Maps API, thus we do not anticipate any potential negative social impact arising from this work. 

\end{document}